\ifdefined\XeTeXversion\else\pdfoutput=1\fi
\documentclass{article}

\PassOptionsToPackage{table}{xcolor}

\PassOptionsToPackage{numbers,compress}{natbib}
\usepackage[main,final]{neurips_2026}

\usepackage[utf8]{inputenc}
\usepackage[T1]{fontenc}
\usepackage{graphicx}
\usepackage{hyperref}
\usepackage{url}
\usepackage{fontawesome5}  
\usepackage{booktabs}
\usepackage{amsfonts}
\usepackage{nicefrac}
\usepackage{microtype}

\graphicspath{{./}}

\usepackage{amsmath}
\usepackage{amssymb}
\usepackage{array}
\usepackage{multirow}
\usepackage{tcolorbox}
\usepackage{wrapfig}
\usepackage[table]{xcolor}
\usepackage{xspace}
\usepackage[capitalize,noabbrev]{cleveref}

\definecolor{lightgreen}{HTML}{B6DEC2}
\definecolor{lightred}{HTML}{FCCAC5}
\definecolor{lightblue}{HTML}{C8EAF5}
\definecolor{Highlight}{HTML}{fc8d62}

\newlength\savewidth

\newcolumntype{x}[1]{>{\centering\arraybackslash}p{#1pt}}
\newcolumntype{y}[1]{>{\raggedright\arraybackslash}p{#1pt}}
\newcolumntype{z}[1]{>{\raggedleft\arraybackslash}p{#1pt}}

\newcommand{\methodname}{Diffusion-DRF}

\definecolor{linkblue}{HTML}{2A5DB0}
\newcommand{\resourcelink}[3]{{\hypersetup{urlbordercolor=white}\href{#1}{\textcolor{black}{#2}\,\textcolor{linkblue}{#3}}}}
\title{Diffusion-DRF: Free, Rich, and Differentiable Reward for Video Diffusion Fine-Tuning}

\author{%
  Yifan Wang$^{1,2,*}$ \quad
  Yanyu Li$^{2}$ \quad
  Gordon Guocheng Qian$^{2,\dagger}$ \quad
  Sergey Tulyakov$^{2}$ \\
  {\bf Yun Fu$^{1}$} \quad
  {\bf Anil Kag$^{2}$} \\
  $^{1}$Northeastern University \quad $^{2}$Snap Inc. \\[0.6em]
  {\normalfont\normalsize
   \resourcelink{https://yfwang1999.github.io/Diffusion-DRF/}{\faGlobe}{Project Page}\hspace{3em}%
   \resourcelink{https://github.com/YFWang1999/Diffusion-DRF}{\faGithub}{Code}%
  }
}

\makeatletter
\let\@oldmaketitle\@maketitle
\renewcommand{\@maketitle}{%
  \@oldmaketitle
  \vspace{-1.2em}
  \teaserfigure
  \vspace{1.0em}
}
\makeatother

\newcommand{\teaserfigure}{%
    \includegraphics[width=\textwidth]{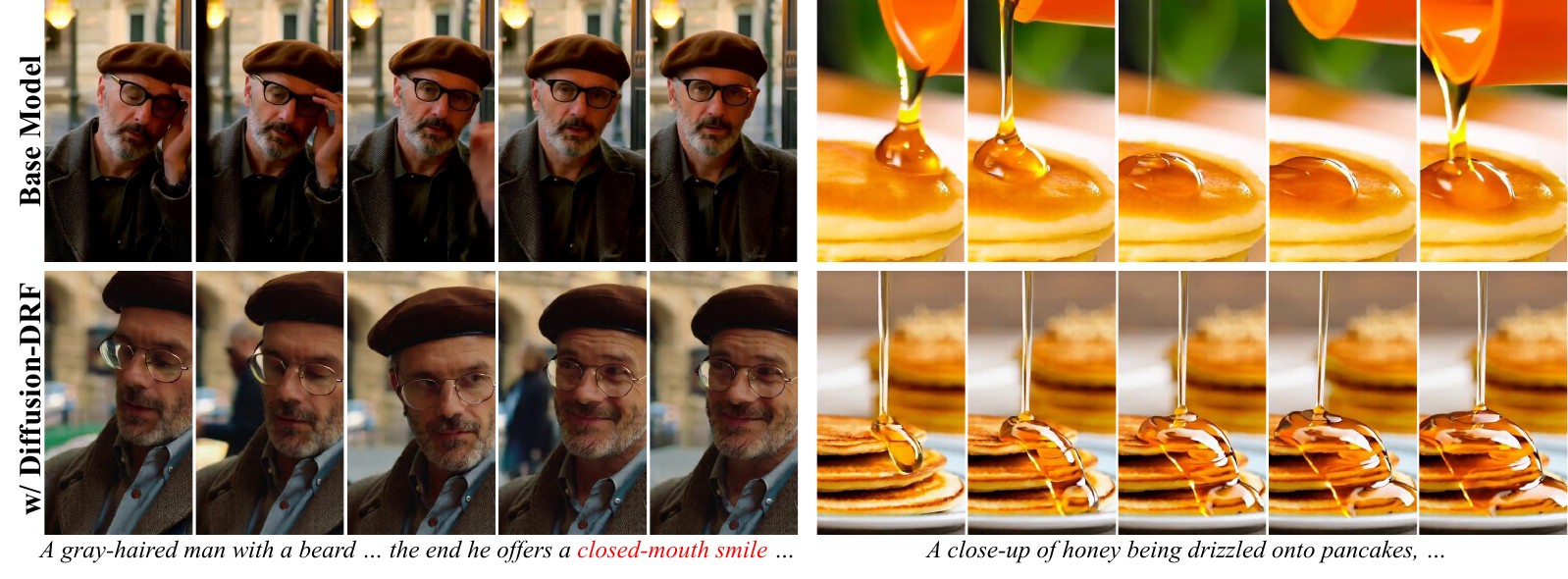}\par
    \vspace{0.35em}
    \refstepcounter{figure}\textbf{Figure \thefigure}. \textbf{Text-to-video fine-tuning with \methodname{}.}
    Our method (bottom row) enhances both text-video alignment (left) and physical fidelity (right), enabling more reliable generation under challenging prompts.
    \label{fig:teaser}
}

\makeatletter\renewcommand{\@notice}{}\makeatother
\begin{document}

\maketitle
\begingroup
\renewcommand\thefootnote{}\footnotetext{$^{*}$Work done during an internship at Snap Inc. $^{\dagger}$ Corresponding author.}
\addtocounter{footnote}{-1}
\endgroup

\begin{abstract}
Video diffusion alignment has been heavily reliant on scalar rewards. These rewards are typically derived from learned reward models in human preference datasets, requiring additional training and extensive collection. Moreover, scalar rewards provide coarse, global supervision, offering limited prompt-generation mismatch credit assignment and making models prone to reward exploitation and unstable optimization.
We propose \textbf{Diffusion-DRF} (Differentiable Reward Flow), a free, rich, and differentiable reward framework for video diffusion fine-tuning, where ``free'' means free of reward-model training and of preference-data collection.
Diffusion-DRF employs a frozen, off-the-shelf Vision-Language Model (VLM) as the critic, eliminating the need for reward model training. Instead of relying on a single scalar reward, it decomposes each user prompt into multi-dimensional questions with freeform dense VQA explanation queries, yielding information-rich feedback. Under teacher forcing, every answer-token loss conditions on the video, question, and a clean target prefix rather than prior model predictions; this non-compounding token supervision prevents answer errors from propagating into subsequent training contexts.
By direct differentiable optimization over this rich feedback, Diffusion-DRF achieves stable reward-based tuning without preference datasets collection. 
Diffusion-DRF achieves significant gains both quantitatively and qualitatively, reaching 55.38 overall on VBench-2.0, 2.39 points above the base model and 1.83 points above the strongest reward-tuning baseline.

\end{abstract}
\section{Introduction}
\label{sec:intro}

 

Recent advances in diffusion-based text-to-video generation~\cite{CogVideoX, HunyuanVideo, Wan2.1, OpenSora, T2V-Turbo-v2VideoDiffusionRewardTuning3,SnapVideo} have markedly improved fidelity, temporal coherence, and prompt adherence. 
Beyond architecture and scaling, a second wave of progress has come from post-training, inspired by alignment practices in LLMs~\cite{DPO, InstructGPT, RLHF-V, shao2024deepseekmath} and text-to-image diffusion \cite{DiffusionDPO, DiffusionRankedDPOSnap, DiffusionMarginDPO, DPOKDiffusionRLTrain2}. 
The core motivation is to 
\begin{wrapfigure}{r}{6.cm}
\centering
\includegraphics[width=\linewidth]{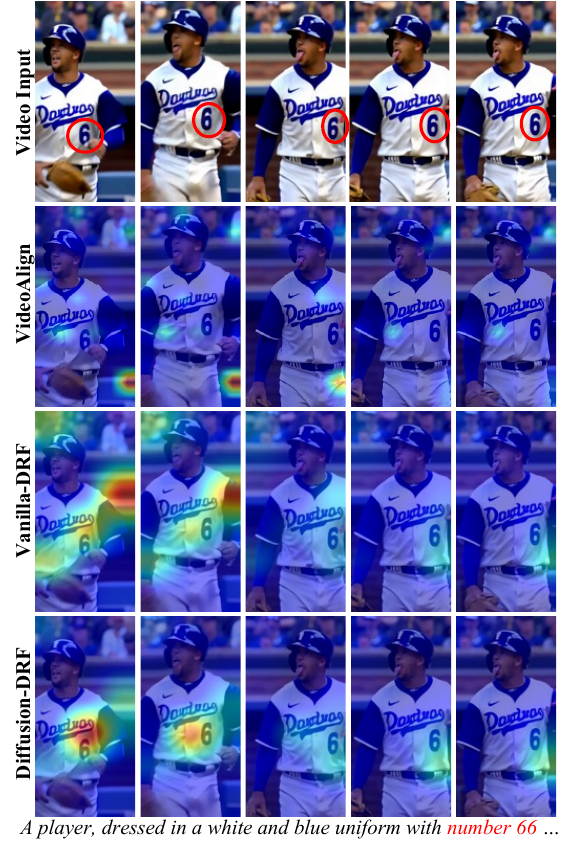}
\vspace{-2em}
\caption{ 
\textbf{Gradient Analysis.}
In contrast to the latest scalar-based VideoAlign~\cite{VideoAlign} and our Vanilla-DRF baseline built on a simple yes/no text-video alignment VLM feedback, 
{our Diffusion-DRF leverages free and rich structured feedback to empirically yield more semantically coherent and spatially concentrated gradients, with heatmaps that better align with regions where the generated video appears to violate the input prompt.} We present more visualizations in \cref{gradient_visualization}.}
\label{fig:grads-intro}
\vspace{-2em}
\end{wrapfigure}
use preference-driven objectives to steer pretrained generators toward human-preferred behaviors that maximum-likelihood training does not capture well. 
Accordingly, a growing set of post-training methods, such as preference optimization (e.g., DPO-style objectives~\cite{wu2025densedpo, VideoAlign, VisionReward}), reinforcement learning with human or AI feedback~\cite{DanceGRPO, Flow-GRPO, SePPO}, and other reward-driven refinements~\cite{VaderVideoDiffusionRewardTuning1, InstructVideoVideoDiffusionRewardTuning2}, have received increasing attention to align model outputs with human judgments. 
Despite these gains, most pipelines still rely on \textit{hand-labeled, scalar} preference signals, either from separate reward models or large-scale DPO-style pairwise annotations.
Such supervision is easily exploited due to its limited scalar supervision and label-intensive, often resulting in reward hacking, training instability, or policy collapse.

Effective supervision should provide \textit{token-level prompt-generation mismatch credit assignment, indicating where and when generated content mismatches the input prompt}. Such structured feedback, rather than a single scalar signal, enables more stable gradient propagation and mitigates reward hacking or training collapse. However, prior reward tuning approaches~\cite{Flow-GRPO,VaderVideoDiffusionRewardTuning1,InstructVideoVideoDiffusionRewardTuning2,VideoAlign} typically produce a single scalar score for an entire video. This global supervision fails to distinguish localized spatial or temporal  text-video inconsistencies, allowing reinforcement learning updates to under-penalize misaligned regions while over-rewarding globally correlated artifacts. This imbalance often leads to reward exploitation and suboptimal optimization dynamics. To better understand the limitation, we conduct a gradient-level analysis of the recent scalar-reward model VideoAlign~\cite{VideoAlign}. Shown in~\cref{fig:grads-intro}, their gradients are spatially diffuse and frequently misplaced, failing to provide precise mismatch credit assignment.

Interestingly, we observe that a pretrained Vision-Language Model (VLM)~\cite{Qwen2.5-VL}, owing to its strong generalization capability, already serves as a powerful visual quality assessor. Even a simple scalar yes or no alignment signal derived from a frozen VLM forms a strong differentiable reward baseline, which we denote as \textbf{Vanilla-DRF}. 
{Despite its simplicity, Vanilla-DRF empirically yields more focused token-level gradients, which in our visualizations better indicate where video content deviates from the prompt.} Empirically, this basic VLM-based reward already surpasses the latest prior method, VideoAlign, by a significant margin (\cref{fig:grads-intro}). However, Vanilla-DRF remains inherently limited by its scalar nature, which can lead to noisy supervision and susceptibility to reward exploitation.

To address these limitations, we propose \textbf{Diffusion-DRF, a free, rich, and Differentiable Reward Flow (DRF) framework} that leverages a frozen, off-the-shelf VLM to fine-tune video diffusion models. We design an automatic prompting pipeline that elicits structured multi-dimensional feedback. Specifically, each prompt is decomposed into multiple targeted questions assessing text-video alignment, visual quality, and physical plausibility, producing binary signals across distinct alignment dimensions, complemented by freeform VQA-style explanations for dense contextual grounding. Together, these responses form \textit{fine-grained, multi-dimensional supervision that is substantially denser and less prone to exploitation than conventional scalar rewards}. Crucially, our teacher-forced next-token objective conditions every answer token on the generated video, its question, and a clean reference-answer prefix rather than previously predicted tokens. We call this property \textit{non-compounding token supervision}: an erroneous token prediction cannot enter the context used to supervise subsequent tokens, avoiding error propagation along the answer sequence during training.
{This formulation empirically provides more precise temporally localized learning signals (Figure \ref{fig:grads-intro} row 4) that improve training stability and mitigate reward hacking (see Sec. \ref{sec:analysis}), while eliminating the need for additional reward model training or large-scale preference annotation, resulting in an extensible alignment pipeline.}

\textbf{Our contributions are summarized as follows:}
\vspace{-0.5em}
\begin{itemize}
\item We propose \textbf{Diffusion-DRF}, a free, rich, and differentiable reward framework for video diffusion fine-tuning. Diffusion-DRF employs a frozen, off-the-shelf VLM as a training-free critic, eliminating the need for reward model training or preference dataset collection.

\item We design an automatic annotation pipeline that decomposes each prompt into multiple targeted questions with binary responses, complemented by freeform VQA explanations. The teacher-forced objective provides non-compounding token supervision, as every token is trained with a clean reference-answer prefix rather than preceding model predictions. This produces multi-dimensional, information-rich feedback that yields temporally and token-aware gradients, substantially improving stability and reducing reward hacking.

\item Extensive experiments demonstrate that Diffusion-DRF achieves stable reward-based tuning under both single-prompt and large-scale settings, consistently improving generation quality and reaching 55.38 overall on VBench-2.0, 2.39 points above the base model and 1.83 points above the strongest reward-tuning baseline, VideoAlign.
\end{itemize}
\section{Related Works}
\label{sec:formatting}

\noindent {\bf Diffusion Model Post-training.}
Recent progress in LLM post-training \cite{ouyang2022instructgpt, bai2022constitutional, rafailov2023direct} has been adapted to visual generation to further improve output quality.
Existing approaches can generally be divided into two paradigms: reinforcement learning–based optimization and differentiable reward optimization.
RL-based methods optimize generators using preference supervision through policy-style updates. 
Among them, Direct Preference Optimization (DPO)~\cite{rafailov2023direct} and Group Relative Policy Optimization (GRPO)~\cite{shao2024deepseekmath} have become popular choices, as they optimize preference signals through policy-style updates. 
These methods have been widely explored in both image \cite{wallace2024diffusion, yang2024using, zhu2025dspo, karthik2025scalable, zheng2025diffusionnft,Flow-GRPO} and video generation \cite{wu2025densedpo, cheng2025realdpo, VideoAlign, VisionReward, zheng2026sagegrpo}.
In parallel, differentiable-based approaches update generators using gradients derived from a reward model, enabling end-to-end optimization through the generation process \cite{VaderVideoDiffusionRewardTuning1, InstructVideoVideoDiffusionRewardTuning2, LiFTVideoDiffusionLossWeighted2, VisionReward, VideoScore}.
Despite notable perceptual improvements, existing approaches typically rely on reward models trained from human preference data, which are annotation-intensive, narrow in coverage, and inherently biased. 
Moreover, these rewards are usually scalar, providing only sparse and coarse supervision. As a consequence, optimization can lead to reward hacking and unstable updates. 
These limitations have motivated follow-up efforts toward richer and more fine-grained feedback~\cite{RichHumanFeedbackGoogle}, but such approaches require even more extensive data collection.


\noindent{\bf Learning from AI Feedback}
Recent advancements have demonstrated that Vision–Language Models (VLMs) can transcend simple image-text matching to perform complex tasks such as aesthetic assessment, fine-grained grounding, and compositional reasoning. 
Thus, VLMs have functioned as automated oracles capable of judging various dimensions of visual quality which are widely used in different T2I/T2V tasks such as data selection~\cite{chen2024sharegpt4video, chen2024panda70mcaptioning70mvideos, nan2024openvid, Dreamsync, wu2024boosting} and evaluation~\cite{VBench, zheng2025vbench, Geneval, hu2023tifa}.
In particular,
\cite{black2023training} uses VLM as a zero-shot reward model to quantify prompt–image alignment within a reinforcement learning framework. \cite{furuta2024improving} leverages Gemini-generated preference data to perform DPO fine-tuning. 
\cite{luo2025dual} uses VLM as a differentiable scalar reward to improve multimodal control.  Despite these advances, directly optimizing video diffusion models with differentiable VLM feedback remains non-trivial and largely unexplored. 

\section{Method}\label{sec:method}

Given a text prompt $c$, a video diffusion model $v_\theta(\mathbf{z}_t \mid c)$ generates a video latent $\mathbf{z}_0$ through iterative denoising from Gaussian noise $\mathbf{z}_T$ in latent space, where $T$ denotes the total number of diffusion steps. 
Our goal is to fine-tune the pretrained model $v_\theta$ using feedback derived from a reward function $f(\mathbf{v}, c)$, where $\mathbf{v}$ denotes the predicted clean video in pixel space decoded from $\mathbf{z}_0$.

\textbf{\methodname{}} leverages a frozen, off-the-shelf Vision-Language Model (VLM) as a training-free critic, without learning a separate reward model or collecting preference pairs. 
We first introduce a minimal formulation, \textbf{Vanilla-DRF}, with a single binary reward in \cref{sec:method:vanilla}. 
We then extend it to our proposed structured multi-dimensional rich reward formulation in \cref{sec:method:diffusion-drf}, and finally introduce our automatic annotation pipeline in \cref{sec:method:data}. 

\subsection{Vanilla-DRF: Single Binary Differentiable Reward}
\label{sec:method:vanilla}

Vanilla-DRF is a minimal formulation of \methodname{}, which also provides free and differentiable VLM feedback. The difference is that Vanilla-DRF relies on a simple single binary reward.

{\noindent\bf Single Binary Reward.}
As a minimal formulation, we query the VLM with a single binary question:
\emph{``Does this video match the prompt?''}
The VLM produces logits $y$ over \texttt{Yes} and \texttt{No}.

{\noindent\bf Logit-level Reward.}
We define a differentiable reward using the log-probability:
\begin{equation}
R_{\text{vanilla}}(\mathbf{v}, c)
=
\log p(y \mid \mathbf{v}, c),
\end{equation}

{\noindent\bf Truncated Backpropagation Through Diffusion.}
Let $\mathbf{z}_T \rightarrow \cdots \rightarrow \mathbf{z}_0$ denote the full denoising trajectory in training. 
We truncate gradients at step $T-K$ by applying a stop-gradient operator to $\mathbf{z}_{T-K}$, and backpropagate only through the final $K$ updates:

\begin{equation}\label{eqn:backprop}
\nabla_\theta R_{\text{vanilla}}
=
\frac{\partial R}{\partial \mathbf{v}}
\frac{\partial \mathbf{v}}{\partial \mathbf{z}_0}
\sum_{k=1}^{K}
\frac{\partial \mathbf{z}_{k-1}}{\partial \theta},
\qquad
\mathbf{z}_{K} \leftarrow \mathrm{sg}(\mathbf{z}_{K}),
\end{equation}

Here $\mathrm{sg}(\cdot)$ denotes the stop-gradient operator, which treats $\mathbf{z}_{T-K}$ as constant with respect to $\theta$. 
As a result, gradients flow only through the last $K$ denoising steps for optimized memory efficiency.

\begin{figure*}[t]
\vspace{-2mm}
\centering
\includegraphics[width=\linewidth]{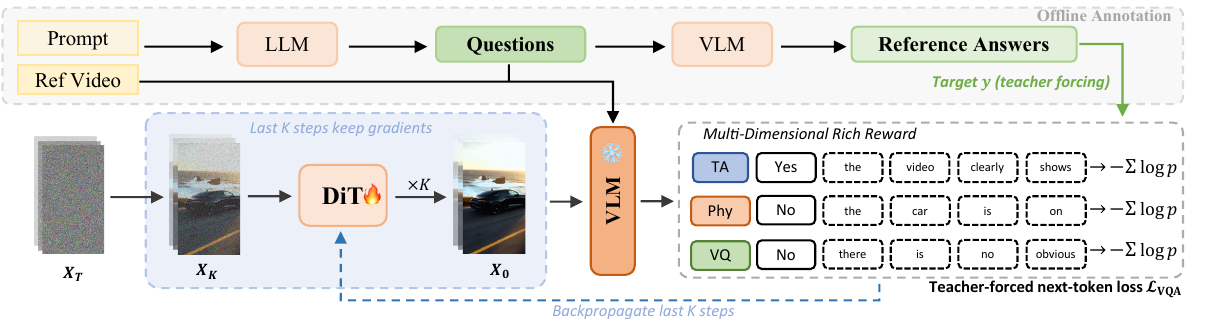}
\vspace{-1.5em}
\caption{
\textbf{Pipeline of \methodname{}.}
Given a reference video and its caption as the prompt, our \methodname{} framework queries a frozen VLM with structured multi-dimensional questions spanning text-video alignment (TA), physical fidelity (Phy), and visual quality (VQ). 
For each dimension, the VLM produces a Yes/No binary judgment together with a dense free-form explanation. 
The logit difference through next-token prediction defines a differentiable reward for each query, and the aggregated multi-dimensional rewards are backpropagated through the VAE decoder and the final $K$ denoising steps to update the latent diffusion model (e.g., DiT~\cite{DiT}).
}
\label{fig:framework}
\vspace{-2mm}
\end{figure*}
\subsection{\methodname{}: Multi-Dimensional Rich Differentiable Reward}
\label{sec:method:diffusion-drf}

The single binary judgment in Vanilla-DRF compresses entity presence, motion correctness, and interaction consistency into one sparse scalar that entangles multiple failure modes: the model can raise the overall \texttt{Yes} score by exploiting superficial cues while receiving little guidance on which component of the generation should be corrected, often leading to unstable or misaligned updates (\cref{gradient_visualization}). 
To obtain structured and harder-to-exploit supervision, we extend the formulation to multi-dimensional rich feedback formed by a set of targeted binary queries derived from explicit semantic and physical decompositions with their corresponding freeform dense explanations. 

{\noindent\bf Structured Query Set.}
Given a prompt $c$, we automatically construct a set of sub-questions for distinct dimension, $\mathcal{Q}(c)=\{q_i(c)\}_{i=1}^N$, where each $q_i$ probes a specific and localized aspect of the generation. 
The query set is organized along three complementary dimensions in our implementation:

\begin{itemize}
    \item \textit{Text-Video Alignment (TA):} semantic entities, attributes, actions, and interactions explicitly described in $c$;
    \item \textit{Physical Fidelity (Phy):} motion realism and physical plausibility;
    \item \textit{Visual Quality (VQ):} perceptual fidelity including sharpness, artifacts, texture preservation, and exposure stability.
\end{itemize}

Each query $q_i$ is instantiated as a minimal Yes/No question followed by an extra query of \textit{free-form explanation}. 
The VLM is prompted to first produce a binary decision token and then generate a dense textual rationale describing where and why the video succeeds or fails for that specific dimension. 
Because the rationale follows the decision, we treat it as \emph{rationale supervision} rather than chain-of-thought reasoning: its tokens provide additional, video-dependent likelihood terms, and placing the explanation before the answer performs equally well (\cref{sec:explanation_ablation}). 


\begin{figure*}[t]
\centering
\includegraphics[width=\linewidth]{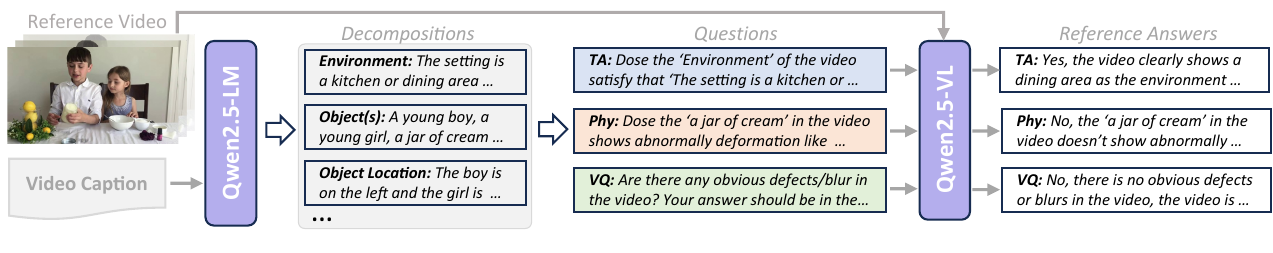}
\vspace{-6mm}
\caption{
\textbf{Automatic Annotation Pipeline.}
Instead of relying on vague global questions or human preference pairs, we construct structured queries directly from captioned videos. 
Given a video caption as the prompt, we extract atomic key points and formulate targeted questions across three domains: text-video alignment (TA), physical fidelity (Phy), and visual quality (VQ). 
Each question is posed in a minimal and unambiguous Yes/No format with a constrained free-form explanation, encouraging precise and localized judgments from the frozen VLM. 
For Phy and VQ, the same queries are additionally applied to caption-matched real videos to obtain reference answers that anchor physical and perceptual assessment. 
Detailed prompts and domain facets are provided in \ref{sec:detailed_prompts}. 
 }
\label{fig:ta-pipeline}
\vspace{-10pt}
\end{figure*}

{\noindent\bf Reference-Guided Queries.}
For physical fidelity (Phy) and visual quality (VQ), we adopt a reference-guided evaluation mechanism. 
Given a caption-matched real video $\mathbf{v}^{\text{ref}}$, the VLM evaluates the generated video $\mathbf{v}$ contrastively, $f(\mathbf{v}, \mathbf{v}^{\text{ref}}, q_i)$, where the reference clip provides a physical and perceptual anchor. 
This stabilizes assessment of motion realism, interaction plausibility, and appearance statistics by grounding the judgment in realistic dynamics and textures. 

{\noindent\bf Free, Rich, and Differentiable Rewards.}
For each query $q_i$, the VLM generates a sequence offline, $y_i = \big(y_i^{0}, y_i^{1}, \dots, y_i^{(L)}\big)$, where the first token $y_i^{0} \in \{\texttt{Yes}, \texttt{No}\}$ represents the binary judgment, and the remaining tokens form a dense free-form explanation, where $L=512$ represents the maximum number of response tokens. This constrained format enforces localized judgments while preserving rich diagnostic information.

{\noindent\bf \methodname{} Training Objective} is defined as the mean next-token negative log-likelihood over all valid target tokens:
\begin{equation}
\mathcal{L}_{\text{VQA}}
=
-\frac{1}{\sum_{i,t} m_{i,t}}
\sum_{i=1}^{N}
\sum_{t=1}^{L_i}
m_{i,t}\,
\log p_\phi
\!\left(
y_i^{(t)}
\mid
\mathbf{c}_i, y_i^{(<t)}
\right),
\end{equation}
where the conditioning context $\mathbf{c}_i$ is defined as
\[
\mathbf{c}_i =
\begin{cases}
(\mathbf{v}, q_i), & q_i \in \text{TA}, \\
(\mathbf{v}, \mathbf{v}^{\text{ref}}, q_i), & q_i \in \{\text{Phy}, \text{VQ}\},
\end{cases}
\]
$p_\phi$ denotes the frozen VLM next-token distribution, and $m_{i,t}\in\{0,1\}$ masks padding tokens. Per-query and per-dimension-balanced normalizations perform equally well (\cref{sec:loss_normalization}).

{\noindent\bf Non-Compounding Token Supervision.}
The prefix $y_i^{(<t)}$ in the objective is the fixed reference-answer prefix supplied under teacher forcing, not a sequence sampled from the VLM. Consequently, every token loss is conditioned on the generated video, its structured question (and the reference video for Phy and VQ), and a clean target prefix. An erroneous prediction at an earlier position therefore cannot enter the conditioning context of later positions or compound along the answer sequence. We refer to this training-time property as \emph{non-compounding token supervision}. It removes autoregressive error propagation within the supervised answer sequence while retaining the contextual dependencies encoded by its reference-answer prefix.

This objective aggregates token-level supervision across the TA, Phy, and VQ queries: jointly optimizing the binary decisions and their explanations yields structured, localized gradients for each semantic and physical facet, which improves credit assignment and mitigates shortcut exploitation. 
\textit{Overall, \methodname{} is training-free with respect to reward modeling, provides rich multi-dimensional feedback, and remains fully differentiable through the VLM next-token objective.}

\subsection{Automatic Annotation Pipeline}
\label{sec:method:data}

To enable the multi-dimensional supervision described above, we require structured annotations that decompose each video data into explicit semantic and physical facets. 
We construct such annotations automatically through a lightweight parsing and prompting pipeline, as illustrated in~\cref{fig:ta-pipeline}.

Shown in~\cref{fig:ta-pipeline} ``Decompositions'', given a video caption as the prompt, we first decompose it into atomic scene elements in text, including background environment, objects, humans, attributes, spatial relations, and their interactions. 
This decomposition identifies the explicit entities and behaviors that should appear in the video. 
Concretely, a text-only LLM (Qwen2.5-7B) is prompted with the instruction template in \cref{fig:prompt_decompose_p1,fig:prompt_decompose_p2} to list these elements, and a second instruction turns them into questions using the dimension-specific templates in \cref{fig:ta_prompts,fig:phy_prompts,fig:vq_prompts}.
The extracted components are then incorporated into the formulation of one structured Yes/No question for each supervision dimension shown in ``Questions''. Each dimension therefore produces one Yes/No judgment accompanied by a dense free-form explanation that references the decomposed components.
The decomposed entities and interactions are also embedded directly into both the question wording and the expected explanations (\cref{fig:ta-pipeline} ``Reference Answers''), ensuring that the VLM grounds its judgment in concrete semantic and physical details.

Reference answers are generated once with Qwen2.5-VL-7B under effectively greedy decoding (\cref{sec:reference_targets}): the paired real video is answered for Phy and VQ, and only samples whose reference answers are all \texttt{Yes} are kept (\cref{data_sampling_procedure}). 
The resulting training set contains 5,077 caption--video pairs and 24,252 queries, with a TA:Phy:VQ ratio of roughly 2.5:1:1. 
The three dimensions are operational supervision groups rather than a complete factorization of video quality; further groups, such as long-range temporal consistency, can be added without changing the objective. 
This automated annotation pipeline replaces vague global evaluation with rich, grounded, aspect-aware supervision. 
It requires captioned real-video pairs for reference-guided training, avoids reward model training and human preference collection, and scales naturally to large datasets. 
By injecting explicit semantic and physical elements into both the question and explanation, the resulting feedback becomes more informative, interpretable, and stable for diffusion fine-tuning.

\section{Experiments}\label{sec:exp}

\subsection{Experimental Setup}\label{sec:exp:setup}

{\noindent\bf Training Details.} We apply \methodname{} to the pretrained Wan2.1-1.3B-T2V~\cite{Wan2.1} with Qwen2.5-VL-7B~\cite{Qwen2.5-VL} as the default VLM. We only train the DiT and freeze other components (VAE, text-encoder, and VLM). 
Videos are generated at $512 \times 288$ resolution with $49$ frames under $25$ denoise steps during training and $30$ steps at inference. 
AdamW~\cite{AdamW,Adam} optimizer is used with a learning rate of $1e^{-5}$, $16$ A100 80GB GPUs, and batch size $1$ per GPU. The training is conducted on $5$K high-quality caption--video pairs from OpenVid-1M~\cite{nan2024openvid} (see \cref{data_sampling_procedure,sec:data_statistics}): each caption prompts an on-policy sample from the generator, and its paired real video serves only as the Phy/VQ reference. Real videos are never used as reconstruction targets, and inference is reference-free. The automatic annotation pipeline is processed offline by Qwen2.5-7B.

{\noindent\bf Efficiency Optimization.}
Back-propagating VLM feedback through the full diffusion chain is prohibitively memory-intensive, as both the DiT denoiser and the frozen VLM are multi-billion parameter models with large activation footprints. A naive implementation quickly exceeds GPU VRAM limits. To make differentiable reward flow practically feasible, we design a set of optimizations:
\emph{1) Efficient Video Decoding.}
Instead of using the original heavy decoder, we trained an efficient H3AE VAE decoder following~\cite{wu2025h3ae} that reproduces the decoding quality of the original decoder with a lighter backbone and employ it only in the reward path, achieving $24.3\%$ less memory consumption in decoding; swapping it for the original decoder changes VBench-2.0 Overall by only 0.3 points (\cref{sec:h3ae}). 
\emph{2) Frame Subsampling.}  To further control VLM activation growth with video length, we uniformly subsample frames before feeding them to the VLM. In practice, we sample $10$ frames from the generated video and $10$ reference frames from a caption-matched real video, which preserves semantic coverage while substantially reducing memory usage (see \cref{sec:frames_ablation} for ablation).
\emph{3) Denoising Aware Activation Checkpointing.}
We apply gradient checkpointing across diffusion steps to trade additional computation for lower memory usage. Concretely, we retain only the input latent at each denoising step and re-compute intermediate DiT activations during the backward pass. This avoids storing the full denoising trajectory in memory, enabling stable end-to-end gradient propagation through the diffusion chain.
\emph{4) Truncated Back-propagation.} While checkpointing allows backpropagation through the entire sampling chain, we find that truncating the backward pass to the last $K$ sampling steps (see~\cref{eqn:backprop}) substantially improves optimization speed and overall performance \cite{DRaFTDiffusionRewardTuning1, huang2025selfforcingbridgingtraintest}. This yields a biased but empirically effective gradient estimator, whose performance--memory trade-off we ablate in \cref{sec:truncation_ablation}.

\begin{table*}[t]
    \vspace{-1mm}
    \centering
    \setlength{\tabcolsep}{3.5pt}
    \scriptsize
\caption{
\textbf{Quantitative results on VBench-2.0.}
In addition to the overall score, we report representative sub-dimension metrics closely related to text-video alignment and physical fidelity. 
The best results are highlighted in \textbf{bold}, and the second best are underlined. 
For completeness, we evaluate \methodname{} with three different VLM backbones: InternVL3-1B~\cite{zhu2025internvl3exploringadvancedtraining}, Qwen2.5-VL-3B, and Qwen2.5-VL-7B~\cite{Qwen2.5-VL}, denoted as \methodname{} (1B), (3B), and (7B), respectively. All methods are trained with Wan2.1-1.3B-T2V as base model on the same datasets for fair comparisons. 
}
    \resizebox{1\textwidth}{!}{
    \begin{tabular}{l|c|ccccc|ccccc}
        \toprule
        \multirow{3}{*}{\textbf{Method}}  & \multicolumn{11}{c}{\textit{VBench-2.0}} \\
        \cmidrule{2-12}
         &\multirow{2}{*}{Overall}& \multirow{2}{*}{Creativity} & 
         {Common} & \multirow{2}{*}{Controllability} & Human  & \multirow{2}{*}{Physics} & \multirow{2}{*}{Material}  & Dynamic  & Motion  & Complex  & Cameras \\
         & & &Sense & & Fidelity& & & Attribute &Rationality & Landscape & Motion \\
        \midrule
        Wan2.1-1.3B-T2V~\cite{Wan2.1}  & {52.99}  & {53.79}      & {55.52} & {26.59}       & {80.65}       & {48.40} &36.23&37.00&37.36&17.33&20.68  \\
        Flow-DPO~\cite{VideoAlign} & 50.27 & 
        46.77 & 50.28 &\underline{27.76} & 71.76 & 54.78 & 69.24 & 40.29 & 35.06 & \underline{18.44} & 22.24\\
        
        Flow-GRPO~\cite{Flow-GRPO} &50.64& 44.71&50.85&25.48&77.80&54.37&69.07&36.63&36.21&14.89&19.32  \\
        
        PickScore~\cite{Pick-a-Pic}& 49.62 & 35.97 & 55.23 & 23.88 &{\bf 81.89}&51.13 & 66.67&39.19&36.78&16.89&22.84 \\
        VideoAlign~\cite{VideoAlign} & 53.55 & 49.64 &56.09&{26.32}&\underline{81.43}&54.26&65.26&41.76&37.93&\underline{18.44}&{23.15} \\
        \midrule
        
        
        
        Vanilla-DRF (7B) & 53.70 & 52.70 & 54.66 & 26.64 & 78.55 &55.61 & 65.31 & 36.63 & 36.21 & 16.67 & 19.44 \\
        
        \methodname{} (1B)&{53.10} & {50.85}  & {56.90} & 24.33 & 77.31 &\underline{56.21} &{68.63}&{40.30}&{40.20}&{15.11}&{\bf 28.40} \\
        
        \methodname{} (3B)& \underline{53.72} & \underline{53.93}  & {\bf 57.53} & 25.95 & 76.74 &{54.42} &\underline{70.00}&\underline{42.49}&{\bf 41.38}&{18.22}&{23.15} \\
        
         {\bf \methodname{} (7B)}& {\bf 55.38} & {\bf 54.58} & \underline{56.96} & {\bf 27.98} & 80.51 & {\bf 56.85} & {\bf 75.82} & {\bf 42.86} & \underline{40.23} & {\bf 21.56} & \underline{24.69} \\

         
        \bottomrule
    \end{tabular}}
    \vspace{-1.em}
    \label{tab:quant-vbench2}
\end{table*}

{\noindent\bf Baselines.} 
We compare~\methodname{} to reinforcement learning based methods Flow-DPO~\cite{VideoAlign} and Flow-GRPO~\cite{Flow-GRPO}. For differentiable reward tuning approaches, we compare to PickScore~\cite{Pick-a-Pic} and VideoAlign~\cite{VideoAlign}, which only provide scalar rewards.
All methods fine-tune the same Wan2.1-1.3B-T2V~\cite{Wan2.1} as the base model on the same dataset for fair comparisons.
We evaluate the checkpoints of different training steps and report the best result for each model; full training curves are shown in \cref{fig:training_dyn}, and three independent runs of \methodname{} give $55.44 \pm 0.26$ (\cref{sec:variance}).

{\noindent\bf Evaluation.}
We evaluate text-to-video performance on 
{VBench-2.0~\cite{zheng2025vbench}, which provides $1,013$ prompts covering advanced aspects from human actions to physical phenomena and aggregates $18$ sub-dimensions into five axes: {\it creativity, commonsense, controllability, human fidelity, and physics}. VBench-2.0 is a multi-faceted evaluation suite rather than a single model judge: its sub-dimensions combine generalist evaluators with specialist metrics and are calibrated with human annotations.}
We also use 
{the VideoGen-Eval prompt set~\cite{yang2025videogen}, which includes $400$ instruction-heavy prompts designed to stress text–video alignment with a pairwise evaluation.}

\subsection{Main Results}\label{sec:exp:main}

\begin{figure}[t]
\centering
\includegraphics[width=1.0\linewidth]{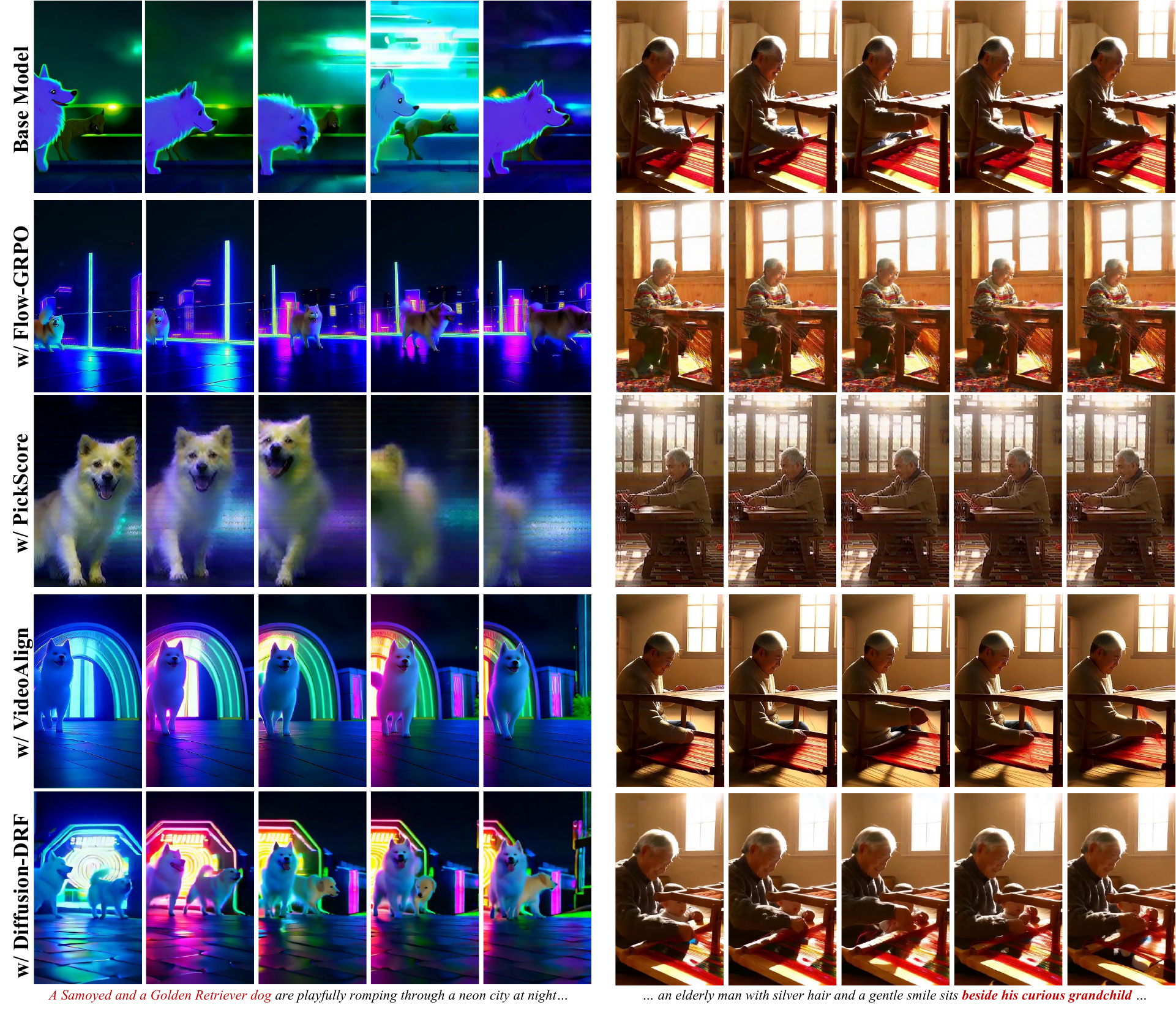}
\vspace{-1em}
\caption{
\textbf{Qualitative Comparison.}
All videos are generated under the same configuration and random seed.
Across both examples, prior reward-tuning baselines either miss key entities or relations in the prompt, or noticeably degrade video quality, while our method better preserves semantic fidelity and visual coherence. For example, competing methods fail to correctly model two different dogs under complex lighting in the left case and omit the child in the right case, whereas our approach faithfully captures both the described motion and entities.
See supplementary videos for full video comparisons across methods.
 }
\label{fig:qual_compare_main}
\end{figure}

\paragraph{Quantitative results on VBench-2.0.}
As shown in Table~\ref{tab:quant-vbench2}, \methodname{} improves the base model across nearly all dimensions with different frozen VLM as reward models. The gains are particularly pronounced on \emph{Dynamic Attribute} and \emph{Physics}, reflecting better adherence to complex prompts and stronger physical plausibility. 
Particularly, scaling the reward model from 1B to 7B yields additional improvements, consistent with the stronger feedback provided by a better VLM. 
Compared to state-of-the-art reinforcement learning tuning methods Flow-DPO~\cite{VideoAlign} and Flow-GRPO~\cite{Flow-GRPO}, our approach consistently improves performance as indicated in ``Overall'' score and achieves much higher quality in ``Creativity'', ``Common Sense'', ``Materials'', ``Dynamic Attributes'', ``Motion Rationality'', and ``Complex Landscape'', demonstrating a better prompt following.

Compared with training dedicated scalar reward models, such as PickScore~\cite{Pick-a-Pic} and VideoAlign~\cite{VideoAlign}, our method leverages a frozen VLM and achieves consistently stronger ``Overall'' performance: 55.38 versus 53.55 for VideoAlign, the strongest baseline, and 52.99 for the base model.
Because \methodname{} keeps improving instead of collapsing, its selected checkpoint is trained longer, so this 1.83-point gain over VideoAlign comes at about 3.2$\times$ its GPU-hours, while the gain over Flow-GRPO comes at a slightly lower cost (\cref{tab:training_cost}).
In practice, these customized reward models are often susceptible to reward hacking, exhibiting biases toward particular visual styles, for example prompt-unrequested stylization (\cref{fig:qual_compare_main,fig:single_prompt_ft}), which ultimately harms creativity and controllability. Consistently, VBench-2.0 Diversity drops from 58.6 for the base model to 29.5 for PickScore and 52.4 for VideoAlign, whereas \methodname{} raises it to 64.4.
Our approach, without reward model training, delivers clear improvements on these dimensions as well as ``Physics'', ``Creativity'', and ``Common Sense'' metrics.


\paragraph{Pairwise Evaluation.}
To complement VBench-2.0, we perform pairwise evaluation on VideoGen-Eval between the base model and \methodname{}.
Table~\ref{human_eval} reports both automatic preference scores from VideoAlign~\cite{VideoAlign}, a reward model trained on human-preference 
\begin{wraptable}{r}{5.8cm}
    \vspace{-1em}
    \caption{{Pairwise evaluation of \methodname{} vs. the base model, including VideoAlign-based automatic preference and human judgments.} 
    }
   \centering
    \setlength{\tabcolsep}{3.5pt}
    \scriptsize
    \resizebox{0.4\textwidth}{!}{
    \begin{tabular}{l|ccc}
        \toprule
        Metrics &  Win (\%) & Tie (\%) & Lose (\%) \\
        \midrule
        VideoAlign & {\bf 53.3} & 11.0 & 35.7 \\
        Human & {\bf 29.0} & 54.0 & 17.0 \\
        \bottomrule
    \end{tabular}}
     \vspace{-1em}
     \label{human_eval}
\end{wraptable}
  data, and blind human judgments on text-video alignment over $200$ randomly sampled prompts, with three matched seeds per prompt and sixteen annotators (9,600 judgments; protocol in \cref{sec:human_protocol}).
Under VideoAlign, \methodname{} wins on $53.3\%$ of pairs and loses on $35.7\%$.
{Human annotators prefer our method in $29.0\%$ of cases versus $17.0\%$ for the base model}, a net preference of $+12.0$ points (95\% CI $[+5.0, +19.0]$, Krippendorff's $\alpha = 0.72$); among decisive comparisons, humans prefer \methodname{} in 63.0\% of cases, close to the 59.9\% under VideoAlign. Because the comparator is the base model, this study establishes improvement over the base model with an evaluator independent of our training signal, not a ranking against the other baselines. 
Additional pairwise comparisons against Flow-GRPO and per-metric breakdowns are provided in \cref{videogen-eval}.

{\noindent\bf Qualitative results.}
Fig.~\ref{fig:qual_compare_main} shows that \methodname{} produces videos with stronger semantic fidelity and visual coherence than the base model, Flow-GRPO~\cite{Flow-GRPO}, VideoAlign~\cite{VideoAlign}, and PickScore~\cite{Pick-a-Pic}, which often miss key entities or degrade visual quality.

\subsection{In-depth Analysis of \methodname{}}
\label{sec:analysis}

\begin{wrapfigure}{r}{6cm}
\centering
\vspace{-1.5em}
\includegraphics[width=\linewidth]{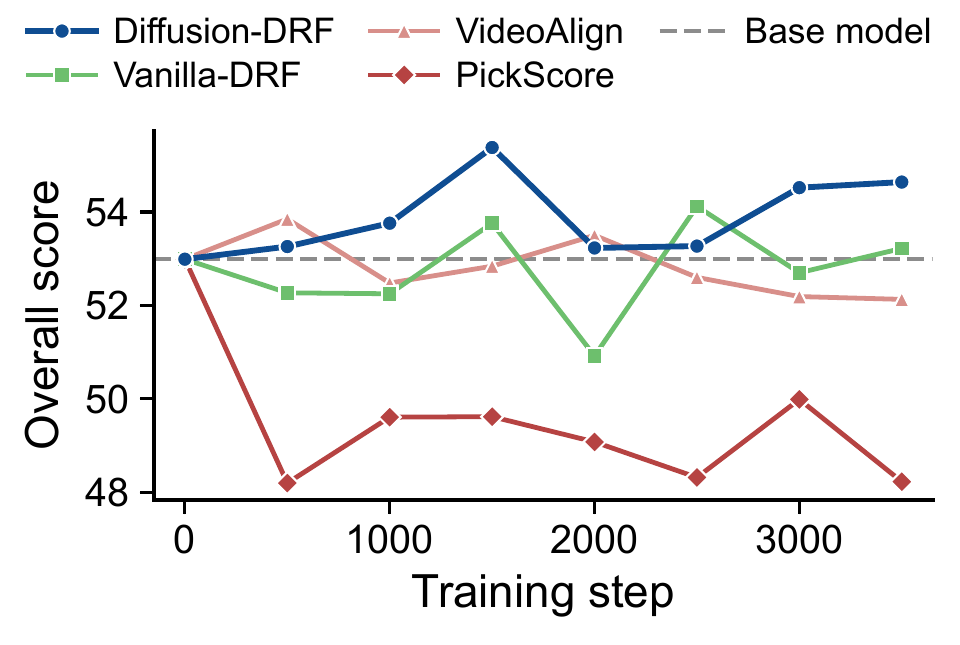}
\caption{
\textbf{Training dynamics} of the same base model optimized with different rewards. 
}
\label{fig:training_dyn}
\vspace{-1em}
\end{wrapfigure}
\paragraph{Training dynamics analysis.}
To examine how different reward formulations affect optimization stability, Fig.~\ref{fig:training_dyn} tracks the VBench-2.0 `Overall' score over an extended training horizon.
\textit{Notably, \methodname{} is the only method whose performance consistently remains above the base model throughout the entire training process. }
In contrast, PickScore~\cite{Pick-a-Pic} deteriorates early and drops far below the base model, indicating clear reward hacking; 
VideoAlign~\cite{VideoAlign} improves in the early phase but gradually declines afterward, again suggesting reward exploitation; 
and Vanilla-DRF, driven by a single scalar reward, shows the same instability.
A likely explanation is that fine-tuning a foundation model into a narrow scalar reward predictor weakens its general video understanding, so it prioritizes superficial cues over the holistic behaviors we aim to improve.
In contrast, \methodname{} keeps the VLM frozen to preserve its general video understanding capacity and leverages multi-dimensional feedback to provide structured supervision that is harder to exploit.

\begin{wrapfigure}{r}{8cm}
\centering
\vspace{-1.5em}
\includegraphics[width=\linewidth]{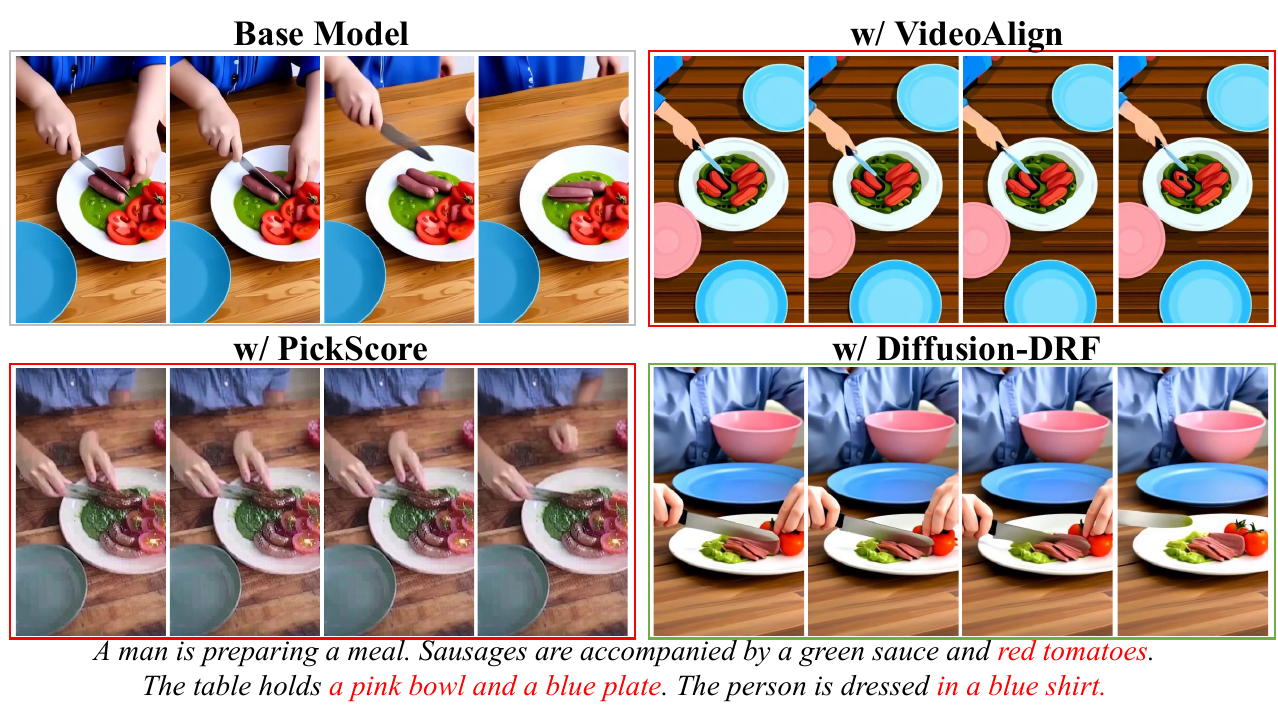}
\vspace{-2.5em}
\caption{\textbf{Qualitative results of single-prompt reward tuning.}
The base model, VideoAlign~\cite{VideoAlign}, and PickScore~\cite{Pick-a-Pic} all exhibit noticeable defects. 
In contrast, Diffusion-DRF produces the most faithful result without introducing visual distortions. 
 }
\label{fig:single_prompt_ft}
\vspace{-1.5em}
\end{wrapfigure}
\paragraph{Single-Prompt Reward Tuning.}
\cref{sec:exp:main} demonstrates that \methodname{} consistently outperforms prior reward tuning approaches. 
To provide a more controlled and intuitive comparison, similar to DreamBooth~\cite{Dreambooth}, which fine-tunes a base model on a single concept, \textit{we introduce a single-prompt reward tuning setting}, where the model parameters are optimized using only one prompt under the guidance of a reward model for 100 steps.
This controlled setting allows us to more directly examine whether a reward model can provide effective supervision.
Shown in Fig.~\ref{fig:single_prompt_ft}, when the base model fails to generate a satisfactory video, learned reward models such as PickScore~\cite{Pick-a-Pic} and VideoAlign~\cite{VideoAlign} fail to deliver informative signals about where and how the generation is incorrect, leading to clear overfitting: the generated videos show significant style shift or noticeable visual distortions, where the text-video alignment remains low.

\subsection{Ablation Study}
\paragraph{Effect of multi-dimensional feedback.}
We study the effect of multi-dimensional feedback varying the question set used for training. All single-dimension settings (TA only, VQ only, and Phy only) underperform the baseline, suggesting that any isolated objective is easy to exploit.
Once two dimensions are combined, all variants surpass the baseline, and TA+VQ is the strongest pair, indicating that explicit visual-quality supervision helps stabilize alignment-oriented optimization.
Using all three dimensions yields the best result by a clear margin, improving the overall score from $52.99$ to $55.38$.
These results support our core design choice: multi-dimensional questions provide complementary constraints that reduce reward hacking and lead to more balanced optimization.
We also provide the detailed results in \cref{sec:question_set_ablation}.

\begin{wraptable}{r}{4.8cm}
    \vspace{-0.5em}
    \centering
    \setlength{\tabcolsep}{6.5pt}
    \scriptsize
    \caption{
        \textbf{Ablation studies on question sets and styles.} 
    }
    \resizebox{0.3\textwidth}{!}{
    \begin{tabular}{l|c}
        \toprule
        {Question sets}  & {\textit{VBench-2.0}} \\
        \midrule

        Baseline & 52.99 \\

        TA Only & 52.15  \\
        VQ Only & 52.45 \\
        Phy Only & 51.81 \\
        TA + Phy & 53.55  \\

        TA + VQ  & \underline{53.92}  \\

        VQ + Phy &  53.20 \\

        TA + VQ + Phy & {\bf 55.38}   \\

        \midrule
        Binary question & 53.30  \\

        Explanation w/ CoT  & 54.85   \\
        \bottomrule
    \end{tabular}}
    \label{tab:quant-ablation}
\end{wraptable}
\paragraph{Effect of explanation.} We study the role of free-form explanations through two ablations: (1) removing the explanation and keeping only binary questions; and (2) varying the length of the reference explanations used for supervision.
When only binary questions are used, the feedback becomes significantly less informative, resulting in weaker optimization and degraded performance, as shown in \cref{tab:quant-ablation} ``Binary question''.
This confirms that binary signals alone provide limited guidance for correcting fine-grained errors. 
The gain comes from the content of the explanations rather than from their length: a length-matched control that pairs each query with a shuffled explanation scores 52.47, below the answer-only objective (\cref{sec:explanation_ablation}).
We further analyze the impact of explanation length. By default, we use reference explanations with an average length of 50 tokens. We then adopt Chain-of-Thought (CoT) prompting to generate detailed reference explanations (denoted as ``Explanation w/ CoT'' in the table). CoT-style explanations help early in training (54.85 at 2K steps) but collapse earlier than concise explanations (51.89 at 3K steps), which we attribute to verbose, weakly grounded tokens diluting the gradient; see \cref{sec:explanation_ablation}.
\section{Conclusion}
\label{sec:conclusion}
We introduced \textbf{\methodname{}}, a video diffusion post-training framework that uses rich, differentiable feedback from a frozen VLM.
Instead of training a separate scalar reward model, our method extracts logit-level signals for text-video alignment, physical fidelity, and visual quality, while free-form explanations provide additional corrective supervision.
With efficient video decoding, frame subsampling, denoising-aware activation checkpointing, and truncated backpropagation, \methodname{} makes such optimization practical for large video generators.
Extensive experiments show consistent gains in prompt adherence, physical plausibility, and perceptual quality, while reducing reward hacking without requiring reward-model training or large-scale preference data.

\paragraph{Limitations.}
The Phy and VQ dimensions require a caption-matched real video for every training prompt, and the gain is contingent on matched references: without them, or with shuffled ones, the score is within run-to-run variance of the base model (\cref{sec:reference_role}).
Our critic and evaluators are both learned models; the human study provides independent evidence only against the base model.
The objective scores one cached wording of each explanation, truncated back-propagation and the H3AE decoder are approximations whose effect we characterize only empirically, the critic sees 10 frames of 49-frame clips, and we test critics only up to 7B parameters.
Finally, \methodname{} still collapses beyond about 3K steps, and its longer stable training costs more GPU-hours than VideoAlign.
We discuss these limitations and future directions in detail in \cref{sec:limitations_future}.


\bibliographystyle{plainnat}
\bibliography{main}

\clearpage
\appendix

\section{Overview}
This appendix provides additional training and inference details (\cref{sec:training_detail}), including the data sampling procedure, training-set statistics, the decoding settings used to generate reference targets, and the training cost of all online methods;
a qualitative and quantitative analysis of reward gradients (\cref{sec:gradient_analysis});
additional quantitative results, the human-evaluation protocol, and run-to-run variance (\cref{sec:additional_results});
additional ablation studies covering question sets, input frames, gradient truncation, explanations, reference videos, loss normalization, and the efficient decoder (\cref{sec:additional_ablations_app});
an expanded discussion of limitations (\cref{sec:limitations_future}) and broader impacts (\cref{sec:broader_impacts});
further qualitative evidence, including single-prompt fine-tuning samples and gradient heatmaps (\cref{sec:visual_results});
and the full prompting configurations (\cref{sec:detailed_prompts}).

\section{Training Details}
\label{sec:training_detail}

\subsection{Configuration}
We provide further training and inference details in this section to support reproducibility.
Overall, we follow the standard configuration of Wan2.1-1.3B-T2V~\cite{Wan2.1}; \cref{tab:wan_config_class} lists the VAE, text-encoder, and scheduler settings.
\begin{table}[h]
    \centering
    \caption{Training Configure.}
    \scalebox{0.9}{
    \begin{tabular}{l|l|c}
        \textbf{class} & \textbf{config} & \textbf{value} \\
        \midrule
        \multirow{2}{*}{VAE} & temporal\_compression\_ratio & 4 \\
         & spatial\_compression\_ratio & 8 \\
        \hline
         \multirow{11}{*}{Text Encoder} 
          & tokenizer & google/umt5-xxl \\
          & text\_length & 512 \\
          & vocab size & 256384 \\
          & dim & 4096 \\
          & dim\_attn & 4096 \\
          & dim\_ffn & 10240 \\
          & num\_heads & 64 \\
          & num\_layers & 24 \\
          & num\_buckets & 32 \\
          & shared\_pos & False \\
          & dropout & 0.0 \\
        \hline
         \multirow{7}{*}{Scheduler} & num\_train\_timesteps & 10000 \\
          & shift & 5.0 \\
          & use\_dynamic\_shifting & false \\
          & base\_shift & 0.5 \\
          & max\_shift & 1.15 \\
          & base\_image\_seq\_len & 256 \\
          & max\_image\_seq\_len & 4096 \\
    \end{tabular}}
    
    \label{tab:wan_config_class}
\end{table}

\subsection{Data Sampling Procedure}
\label{data_sampling_procedure}
We randomly sampled 25k video-caption pairs based on the category distribution in OpenVid-1M~\cite{nan2024openvid}.
These pairs were then processed by our prompting pipeline to get the reference answers.
To ensure the VLM could correctly understand the video, we retained only the cases where all references were `yes'.
This process resulted in our final dataset.

\subsection{Training-Set Statistics and Supervision Scope}
\label{sec:data_statistics}
The final training set contains 5,077 caption--video pairs and 24,252 structured queries, with a TA:Phy:VQ ratio of roughly 2.5:1:1.
Each OpenVid caption is used as the prompt for an on-policy sample from the generator, and its paired real video serves \emph{only} as the reference anchor for the Phy and VQ queries during training.
TA queries evaluate the generated video alone, real videos are never used as pixel- or flow-matching targets, and inference is entirely reference-free.
The data requirement of \methodname{} is therefore \emph{captioned real-video pairs for reference-guided training}; captions alone are sufficient for the TA dimension but not for Phy and VQ (see \cref{sec:reference_role} for the corresponding ablations).
We verified that no training caption appears verbatim in the VBench-2.0 or VideoGen-Eval prompt sets.
Finally, ``free'' in the name of the method means free of reward-model training and of preference-data collection; it does not mean free of computation (\cref{sec:training_cost}) or of training data.
The three supervision groups (TA, Phy, VQ) are operational groupings of the questions that our pipeline produces rather than a unique or complete factorization of video quality; additional groups such as long-range temporal consistency are a natural extension.

\subsection{Reference-Target Generation}
\label{sec:reference_targets}
Reference answers and explanations are generated once, offline, with a fixed Qwen2.5-VL-7B-Instruct checkpoint~\cite{Qwen2.5-VL} using \texttt{do\_sample=True}, \texttt{temperature=0.1}, \texttt{top\_k=1}, \texttt{top\_p=0.001}, \texttt{repetition\_penalty=1.05}, and \texttt{max\_new\_tokens=512}.
Because \texttt{top\_k=1} retains a single candidate per step, decoding is effectively deterministic.
Explanations are constrained to fewer than 50 words by the prompt (\cref{sec:detailed_prompts}), and all targets are cached and reused across methods, ablations, and seeds, so every comparison in the paper is made against identical supervision.

\subsection{Training Cost}
\label{sec:training_cost}
\begin{table}[htbp]
    \centering
    \setlength{\tabcolsep}{6pt}
    \scriptsize
    \caption{
        \textbf{Training cost of online reward-tuning methods.}
        All methods fine-tune Wan2.1-1.3B-T2V on the same 16$\times$A100-80GB node; ``Selected step'' is the checkpoint reported in \cref{tab:quant-vbench2}.
        GPU-hours are computed as $\text{GPUs}\times\text{sec./step}\times\text{selected step}/3600$.
        Flow-DPO is omitted because it trains offline on pre-collected preference pairs at a single sampled timestep and does not roll out videos during optimization.
    }
    \vspace{1mm}
    \begin{tabular}{l|ccccc|c}
        \toprule
        \textbf{Method} & GPUs & Peak VRAM / GPU & Sec. / step & Selected step & GPU-hours & VBench-2.0 \\
        \midrule
        VideoAlign~\cite{VideoAlign} & 16$\times$A100-80GB & 72 GB & 22.50 & 500  & 50.00  & 53.55 \\
        Flow-GRPO~\cite{Flow-GRPO}   & 16$\times$A100-80GB & 68 GB & 75.00 & 500  & 166.67 & 50.64 \\
        \methodname{} (7B)           & 16$\times$A100-80GB & 75 GB & 23.73 & 1500 & 158.17 & {\bf 55.38} \\
        \bottomrule
    \end{tabular}
    \label{tab:training_cost}
\end{table}
\cref{tab:training_cost} reports the GPU configuration, peak per-GPU memory, wall-clock time per optimization step, the selected checkpoint, and the resulting GPU-hours for the three online reward-tuning methods in \cref{tab:quant-vbench2}.
Per-step cost of \methodname{} (23.73 s) is close to that of VideoAlign (22.50 s), since both back-propagate a differentiable reward through the last $K$ denoising steps, and roughly three times lower than that of Flow-GRPO (75.00 s), which must roll out a group of videos per prompt.
The main cost difference comes from the checkpoint selected: because \methodname{} keeps improving instead of collapsing (\cref{fig:training_dyn}), its best checkpoint is at step 1500, whereas VideoAlign peaks at step 500.
Consequently, the 1.83-point gain over VideoAlign in \cref{tab:quant-vbench2} is obtained at about 3.2$\times$ its GPU-hours, while the 4.74-point gain over Flow-GRPO is obtained at slightly lower total cost.
Flow-DPO~\cite{VideoAlign} trains offline on human-annotated preference pairs with a DPO-style flow-matching loss at a single sampled timestep and does not generate rollout videos during optimization, so its cost is not comparable to the online methods above and is omitted.

\section{Gradient Analysis}
\label{sec:gradient_analysis}

\begin{figure}[t]
\centering
\vspace{-2em}
\includegraphics[width=\linewidth]{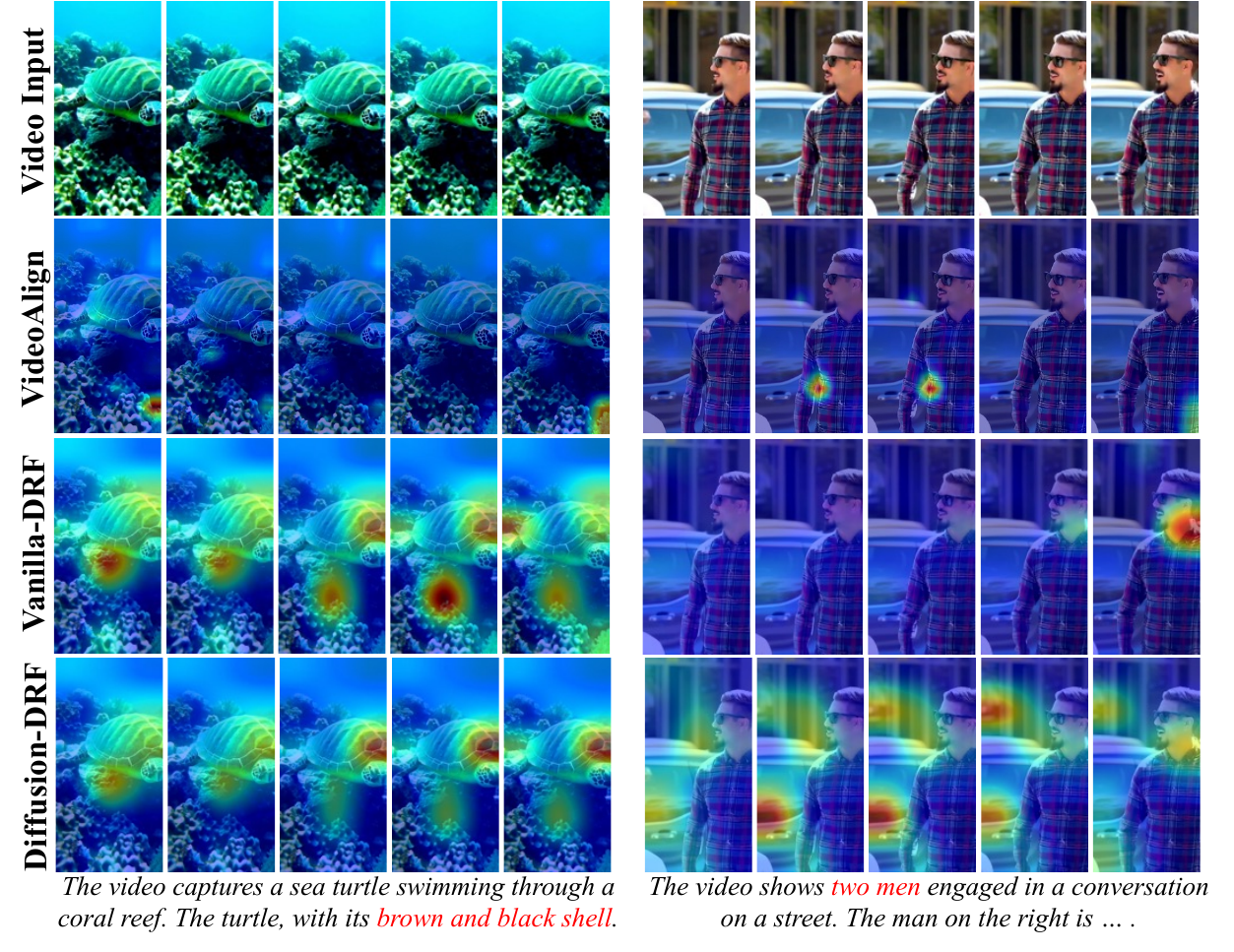}
\vspace{-2em}
\caption{
\textbf{Gradient heatmap visualizations for different reward formulations.}
Original video failed to satisfy the color requirement of the turtle’s shell (left) and two men (right). 
}
\label{fig:grad-visualization}
\end{figure}
\subsection{Gradient Visualization}
\label{gradient_visualization}
Similar to \cref{sec:intro}, we further analyze the gradients induced by different reward formulations.
Gradients directly determine the optimization direction.
The activated regions in a gradient heatmap correspond to the parts of the video that the reward signal considers most responsible for the mismatch with the prompt.
In other words, correctly localized gradient activation indicates that the reward provides semantically aligned corrective signals.
We visualize and compare the gradient distributions produced by customized reward VideoAlign~\cite{VideoAlign}, Vanilla-DRF, and \methodname{}.
As illustrated in \cref{fig:grad-visualization}, when fine-grained details are missing or incorrectly rendered, \textit{only \methodname{} produces spatially localized gradients that precisely highlight the regions requiring modification}.
In contrast, customized reward models~\cite{VideoAlign} trained on narrow preference data often produce diffuse or misplaced gradient signals, reflecting shortcut correlations or dataset bias, compared to ours that use a frozen VLM, a general large-scaled trained critic.
Similarly, without multi-dimensional rich feedback, a single global reward query in Vanilla-DRF can entangle multiple semantic factors, leading to ambiguous optimization signals. As a result, the activated regions do not consistently correspond to the true source of error in the generated video. More visualizations are shown in \cref{fig:additional_heatmaps1,fig:additional_heatmaps2}.

\subsection{Quantitative Gradient Localization}
\label{sec:grad_localization}
\begin{table}[htbp]
    \centering
    \setlength{\tabcolsep}{8pt}
    \scriptsize
    \caption{
        \textbf{Gradient localization on synthetic defects.}
        Localization enrichment (higher is better) is the fraction of absolute reward-gradient mass that falls inside the defect mask divided by the mask's spatio-temporal volume fraction; $1.0\times$ corresponds to a spatially uniform gradient.
        Left: 200 held-out clips with localized blur, noise, or frame-freeze corruptions inserted into foreground or background regions.
        Right: a second set of 200 held-out clips whose captions contain no background description, split by whether the corrupted region is caption-grounded foreground or caption-unmentioned background.
        ``Random masks'' are size-, shape-, and duration-matched control masks.
    }
    \vspace{1mm}
    \begin{tabular}{l|c|cc}
        \toprule
        \multirow{2}{*}{\textbf{Objective}} & \multirow{2}{*}{All defects} & \multicolumn{2}{c}{Split by caption grounding} \\
        \cmidrule{3-4}
        & & Unmentioned background & Caption-grounded foreground \\
        \midrule
        Random masks (control) & 1.0$\times$ & 1.00$\times$ & 1.00$\times$ \\
        Answer only            & 1.3$\times$ & 1.34$\times$ & 1.45$\times$ \\
        TA only                & --          & 1.05$\times$ & 2.34$\times$ \\
        Phy only               & --          & 2.00$\times$ & 2.00$\times$ \\
        VQ only                & --          & 2.15$\times$ & 2.12$\times$ \\
        Full \methodname{}     & {\bf 2.5$\times$} & {\bf 2.23$\times$} & {\bf 2.52$\times$} \\
        \bottomrule
    \end{tabular}
    \label{tab:grad_localization}
\end{table}
The heatmaps above are illustrative, so we additionally quantify how well the reward gradient localizes defects.
The reward is computed from the terminal (fully denoised) video, but a terminal objective can still produce highly non-uniform gradients over pixels and frames; the question is whether the non-uniformity points at the actual errors.

\textbf{Protocol.}
We take 200 held-out clips and insert synthetic defects: localized blur or noise, or a short frame freeze, placed in either a foreground or a background region.
Captions and cached targets are kept fixed, and the clean clip is used as the Phy/VQ reference where applicable.
We back-propagate the reward of each objective to the corrupted video and measure \emph{localization enrichment}: the fraction of absolute gradient mass that falls inside the defect mask divided by the mask's spatio-temporal volume fraction, so that a spatially uniform gradient scores $1.0\times$.
As a control we evaluate the same gradient against random masks matched in size, shape, and duration.
All objectives receive identical inputs.

\textbf{Results.}
Defects occupy about 10\% of the video volume but receive 25\% of the gradient mass of \methodname{}, a $2.5\times$ enrichment, compared to $1.3\times$ for the answer-only objective and $1.0\times$ for random masks (\cref{tab:grad_localization}, left).
This demonstrates content-sensitive localization in decoded-video space; it does not measure credit assignment across denoising steps, which is studied separately through the truncation length in \cref{sec:truncation_ablation}.

\textbf{Content not mentioned in the caption.}
A caption describes only part of a video, so we further ask whether the gradient can localize defects in regions that the prompt never mentions, for example a corrupted background behind ``a cat is walking''.
We repeat the protocol on 200 held-out clips whose captions contain no background description.
Caption-grounded foreground masks are produced with GroundingDINO~\cite{GroundingDINO} and SAM~2~\cite{SAM2}, background masks with panoptic segmentation, both followed by manual verification.
\cref{tab:grad_localization} (right) shows that the TA objective localizes almost exclusively caption-grounded content ($2.34\times$ on foreground versus $1.05\times$ on unmentioned background), whereas the VQ and Phy objectives localize visually or physically defective regions regardless of whether they are named ($2.12\times$/$2.15\times$ and $2.00\times$/$2.00\times$).
Because VQ and Phy assess perceptual quality and physical plausibility rather than pixel correspondence, they can penalize a corrupted background without requiring it to match the reference background; the reference provides calibration rather than a pixel target.
The full objective inherits both behaviors ($2.52\times$ foreground, $2.23\times$ background), which is consistent with the complementary roles of the three dimensions observed in \cref{sec:question_set_ablation}.

\section{Additional Results}
\label{sec:additional_results}

\subsection{Generalization to Another Backbone}
\begin{table}[h]
    \vspace{-2mm}
    \centering
    \setlength{\tabcolsep}{3.8pt}
    \scriptsize
    \caption{
        \textbf{Generalization to CogVideoX.}
        We report VBench-2.0 results of CogVideoX and of CogVideoX fine-tuned with \methodname{}.
        }
    \vspace{1mm}
    \begin{tabular}{l|c|ccccc}
        \toprule
        \multirow{3}{*}{\textbf{Method}}  & \multicolumn{6}{c}{\textit{VBench-2.0}} \\
        \cmidrule{2-7}
        & \multirow{2}{*}{Overall} & \multirow{2}{*}{Creativity} & 
         {Common} & \multirow{2}{*}{Controllability} & Human  & \multirow{2}{*}{Physics} \\
         & & &Sense & & Fidelity& \\
        \midrule
        CogVideoX&49.52& 37.88&54.87&24.56&78.10&52.18 \\
        w/ Diffusion-DRF&51.89 & 40.63&57.75&27.11&81.28&52.69 \\
        
        \bottomrule
    \end{tabular}
    \label{tab:addition_comparison}
    \vspace{-2mm}
\end{table}
We further apply \methodname{} to CogVideoX~\cite{CogVideoX} to evaluate its generalization beyond the main diffusion backbone. As shown in \cref{tab:addition_comparison}, \methodname{} consistently improves the base CogVideoX model across all reported dimensions on VBench-2.0, raising the overall score from 49.52 to 51.89. Notably, the gains are not limited to a single aspect: Creativity improves from 37.88 to 40.63, Common Sense from 54.87 to 57.75, Controllability from 24.56 to 27.11, Human Fidelity from 78.10 to 81.28, and Physics from 52.18 to 52.69. These results suggest that the benefit of \methodname{} is backbone-agnostic, and that the proposed differentiable reward can consistently enhance multiple aspects of video generation rather than over-optimizing a narrow subset of metrics.

\subsection{Additional Pairwise Results on VideoGen-Eval}
\label{videogen-eval}

\begin{figure}[h]
\centering
\includegraphics[width=\linewidth]{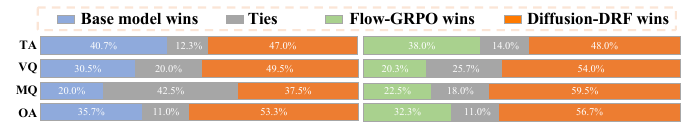}
\caption{
\textbf{Pair-wise evaluation on VideoGen-Eval.}
With the same prompt and configuration, we perform pairwise comparisons of generated videos using the VideoAlign scores of text-video alignment (TA), visual quality (VQ), motion quality (MQ) and overall (OA). 
We compare the Diffusion-DRF with the base model (left) and the Flow-GRPO (right) respectively.
 }
\label{fig:quant_preference}
\end{figure}
\cref{human_eval} in the main paper summarizes the pairwise comparison between \methodname{} and the base model using both VideoAlign-based automatic scoring and independent human judgments.
Here we provide a more detailed automatic analysis with VideoAlign.
For each caption, we generate two videos with the same configuration and compare them using VideoAlign scores, where a score difference under 0.2 is counted as a tie.
\cref{fig:quant_preference} reports win / tie / loss rates for two comparisons: (i) Base vs.\ \methodname{} and (ii) Flow-GRPO vs.\ \methodname{} across text-video alignment, visual quality, motion quality, and overall score.
Across both pairings, \methodname{} attains consistently higher win rates, showing that the gains are not confined to a single metric or baseline.
Because VideoAlign is itself a learned evaluator and also serves as a baseline reward, we treat this analysis as complementary per-metric evidence; the human study below is the independent validation.

\subsection{Human Evaluation Protocol and Statistics}
\label{sec:human_protocol}
The human study in \cref{human_eval} compares \methodname{} against the \emph{base model}, the comparator for which the automatic VideoAlign preference is also reported; it therefore establishes an improvement over the base model with an evaluator that is independent of the training signal, not a ranking against the other reward-tuned baselines.

\textbf{Setup.}
We sample 200 prompts from VideoGen-Eval~\cite{yang2025videogen} and generate three matched-seed pairs per prompt, giving 600 anonymized video pairs.
Sixteen annotators each evaluated all 600 pairs, yielding 9,600 judgments.
Model identities were hidden and the left/right order was randomized per pair.
Annotators were instructed to judge prompt fidelity, namely whether the entities, attributes, counts, actions, interactions, and spatial-temporal relations described in the prompt appear in the video, and to select a tie when neither video was clearly superior.
The win / tie / loss rates in \cref{human_eval} aggregate these judgments.

\textbf{Statistics.}
Humans prefer \methodname{} in 29.0\% of comparisons and the base model in 17.0\%, a net preference of $+12.0$ points with a 95\% confidence interval of $[+5.0, +19.0]$ obtained by clustering on both prompts and annotators.
The large tie rate (54.0\%) reflects that many prompts are already handled well by the base model, in which case \methodname{} preserves the generation (\cref{sec:visual_results}).
Restricted to decisive comparisons, \methodname{} is preferred in 63.0\% of cases (95\% CI $[56.0\%, 70.0\%]$), close to the 59.9\% obtained under VideoAlign ($53.3/(53.3+35.7)$), which argues against the automatic preference being an artifact of evaluator--critic correlation.
Krippendorff's $\alpha$ across annotators is 0.72, indicating substantial inter-annotator agreement.

\textbf{On evaluator independence.}
Our optimization signal comes from Qwen2.5-VL, and both evaluation suites include learned components: VBench-2.0 combines generalist VLM evaluators with specialist metrics calibrated on human annotations, and VideoGen-Eval uses VideoAlign.
Neither shares our critic, but both are learned, so we regard the human study as the primary independent evidence and the automatic metrics as supporting evidence.

\subsection{Run-to-Run Variance}
\label{sec:variance}
\cref{fig:training_dyn} reports complete training trajectories with the same best-checkpoint protocol applied to every method, and \methodname{} remains above all competing methods over a broad interval rather than at a single favorable checkpoint.
To quantify seed variation, we repeat the default \methodname{} (7B) run with three independent seeds and obtain VBench-2.0 Overall scores of 55.38, 55.72, and 55.22, i.e.\ $55.44 \pm 0.26$ (mean $\pm$ standard deviation).
Even the lowest run exceeds VideoAlign (53.55) by 1.67 points and the base model (52.99) by 2.23 points.
We present this as evidence of low run-to-run variation of our method; it is not a significance test against baselines for which only a single run is available, and throughout the appendix we treat differences within about one standard deviation ($\pm0.26$) as not measurable.

\section{Additional Ablation Studies}
\label{sec:additional_ablations_app}
\label{additional_ablations}

\subsection{Question Sets}
\label{sec:question_set_ablation}

\begin{table}[t]
    \centering
    \setlength{\tabcolsep}{5.5pt}
    \scriptsize
    \caption{
        \textbf{Detailed results of ablation studies on question sets and styles.}
    }
    \resizebox{0.95\textwidth}{!}{
    \begin{tabular}{l|c|ccccc}
        \toprule
        \multirow{2}{*}{\textbf{Method}}  & \multicolumn{6}{c}{\textit{VBench-2.0}} \\
        \cmidrule{2-7}
        & Overall & Creativity & Common Sense & Controllability & Human Fidelity & Physics  \\
        \midrule
        Baseline & 52.99 & 53.79 & 55.52 & 26.59 & {\bf 80.65} & 48.40  \\
        TA Only & 52.15 & 49.41 & 53.80 & 27.03 & 74.88 & 55.64  \\
        VQ Only & 52.45 & 50.53 &53.22&27.14&80.22&51.28 \\
        
        Phy Only & 51.81 & 37.04 & 57.82 & 27.14 & 79.46 & \underline{56.28}\\
        
        TA + Phy & 53.55 & 50.56 & {\bf 60.12} & 25.72 & 75.57 & {55.78}  \\

        TA + VQ  & 53.92 & \underline{52.34} & 55.81 & \underline{27.80} & 79.02 & 54.65 \\

        VQ + Phy  & 53.20 & {44.50} & 55.81 & {27.10} & 79.02 & 55.64 \\

        TA + VQ + Phy & {\bf 55.38} & {\bf 54.58} & \underline{56.96} & {\bf 27.98} & \underline{80.51} & {\bf 56.85}  \\

         \midrule
        Binary question & 53.30  & 50.56 & 56.96 & 25.72 & 75.57 & 55.78 \\

        Explanation w/ CoT  & 54.85 & 53.75 & 59.39 & 27.52 & 79.53& 56.72  \\
        \bottomrule
    \end{tabular}}
    \label{tab:quant-ablation-details}
    \vspace{-2mm}
\end{table}







We provide the detailed results of the ablation studies of the question sets in \cref{tab:quant-ablation-details}.
Consistent with the summary in the main paper, all single-dimension settings underperform the baseline, all two-dimension combinations surpass the baseline, TA+VQ is the strongest two-way combination, and TA+VQ+Phy gives the best overall result.
The per-dimension scores further clarify how each feedback type shapes model behavior.
TA focuses on explicit semantic entities, attributes, and actions, so it improves \emph{Controllability} from 26.59 to 27.03 and substantially raises \emph{Physics} from 48.40 to 55.64, but still hurts \emph{Creativity} (49.41), \emph{Common Sense} (53.80), and \emph{Human Fidelity} (74.88), lowering \emph{Overall} from 52.99 to 52.15: TA rewards the requested entities, attributes, and actions but does not explicitly penalize perceptual artifacts or distorted human appearance, so alignment-only supervision is still easy to exploit.
VQ emphasizes perceptual quality, artifacts, and appearance consistency, and therefore best preserves \emph{Human Fidelity} among the single-dimension settings at 80.22 while modestly improving \emph{Controllability} to 27.14; however, its gains on \emph{Creativity} (50.53), \emph{Common Sense} (53.22), and \emph{Physics} (51.28) remain limited.
Phy targets motion realism and physical plausibility, which explains why it gives the strongest single-dimension \emph{Physics} score of 56.28 and also improves \emph{Common Sense} to 57.82, but it severely hurts \emph{Creativity} to 37.04, showing that physical supervision alone can over-regularize the generator.

Combining dimensions leads to much more balanced behavior.
TA+Phy reaches the highest \emph{Common Sense} score in the whole table at 60.12 and maintains strong \emph{Physics} at 55.78, but its \emph{Controllability} (25.72) and \emph{Human Fidelity} (75.57) remain weak.
TA+VQ is the strongest two-way combination in terms of \emph{Overall} score at 53.92, and its gains come from a better balance across \emph{Creativity} (52.34), \emph{Controllability} (27.80), and \emph{Human Fidelity} (79.02), confirming the main-paper observation that visual-quality supervision is particularly important for stabilizing alignment-oriented optimization.
VQ+Phy is also better than the baseline, with an \emph{Overall} score of 53.20, but it remains below TA+VQ, suggesting that physical and perceptual feedback alone are still insufficient without explicit semantic alignment constraints.
Finally, TA+VQ+Phy gives the best \emph{Overall} score of 55.38 and the strongest or near-strongest values on all major axes, including \emph{Creativity} (54.58), \emph{Controllability} (27.98), \emph{Human Fidelity} (80.51), and \emph{Physics} (56.85). This result supports our central claim that the three dimensions provide complementary constraints: TA anchors prompt semantics, VQ protects perceptual quality, and Phy improves motion and physical plausibility.

\subsection{Number of Input Frames}
\label{sec:frames_ablation}
\begin{table}[t]
    \vspace{-2mm}
    \centering
    \setlength{\tabcolsep}{3.8pt}
    \scriptsize
    \caption{
        \textbf{Additional ablation studies on input frames.}
        We report the results of the ablations on changing the number of input frames for the VLM on VBench-2.0. All models are trained with the same configuration as Diffusion-DRF.
        }
    \vspace{1mm}
    \begin{tabular}{l|cccccc|c}
        \toprule
        \multirow{3}{*}{\textbf{Method}}  & \multicolumn{6}{c|}{\textit{VBench-2.0}} & Training \\
        & \multirow{2}{*}{Overall}& \multirow{2}{*}{Creativity} & 
         {Common} & \multirow{2}{*}{Controllability} & Human  & \multirow{2}{*}{Physics} & {VRAM} \\
         & & & Sense & & Fidelity& & Usage (Gb)\\
        \midrule
        $N_f=2$ &50.83&48.43&48.10&27.50&80.96&54.14 & 62\\
        $N_f=4$ &51.45&43.22&53.51&27.64&78.4&54.45 & 65 \\
        $N_f=6$ &52.19 & 43.85&55.23&29.25&78.88&53.75 & 68\\
        $N_f=8$ &54.80 & 54.50&55.81&28.10&79.61&56.03 & 71 \\
        $N_f=10$ & 55.38 & {54.58} & {56.96} & {27.98} & 80.51 & {56.85} & 75 \\
        \midrule
        $K=4$ & 55.38 & {54.58} & {56.96} & {27.98} & 80.51 & {56.85}  & 75\\
        $K=3$ & 55.05 & 52.75 & 58.11& 27.52 &80.80 &56.09 & 69\\
        $K=2$ & 54.36 & 52.45&57.24&27.45&78.50&56.18 & 63\\
        $K=1$ & 53.16 & 52.95&52.93&25.45&79.49&54.99 & 56\\
        \midrule
        CoT (1K-steps) & 54.25 & 52.75 & 58.11 & 26.52 & 79.80& 56.09& 75  \\
        CoT (2K-steps) &54.85 & 53.75 & 59.39 & 27.52 & 79.53& 56.72 & 75 \\
        CoT (3K-steps) & 51.89 & 43.75 & 55.81 & 24.83 & 79.13 & 53.92 & 75 \\
        \midrule
        w/o ref frames & 53.15 & 50.41 & 54.50 & 27.80 & 78.88 & 56.34& 72 \\

        \bottomrule
    \end{tabular}
    \label{tab:additional_ablation}
    \vspace{-2mm}
\end{table}
We study how the number of video frames fed into the VLM affects training.
Generated videos contain 49 frames; $N_f$ denotes the number of uniformly subsampled frames given to the critic (and, for Phy/VQ, to the reference), not the video length.
As shown in \cref{tab:additional_ablation}, increasing $N_f$ leads to a clear and consistent improvement in overall performance, with the score rising from 50.83 at $N_f=2$ to 55.38 at $N_f=10$; the last step from $N_f=8$ to $N_f=10$ still adds 0.58 points (54.80 to 55.38) at the cost of 4 GB of memory (71 GB to 75 GB).
This trend indicates that denser temporal observations help the VLM better understand the video content and provide more informative reward signals. In particular, larger $N_f$ also improves several key dimensions, including Common Sense and Physics, suggesting that richer temporal coverage is especially helpful for reasoning about object dynamics and physical plausibility.
For these 49-frame clips we observe that the critic's responses remain largely unchanged when $N_f$ exceeds 10, suggesting that the feedback saturates at the current duration, so we use $N_f=10$ as the coverage--memory trade-off.
This saturation should not be extrapolated to longer videos, where sparse sampling may miss multi-stage actions or delayed physical effects (\cref{sec:limitations_future}).

\subsection{Gradient Truncation Length}
\label{sec:truncation_ablation}
We set the number of back-propagation steps $K$ to 4 in the main results and ablate its effect in \cref{tab:additional_ablation}.
Back-propagating only through the final $K$ denoising steps yields a biased pathwise gradient estimator of the full-chain gradient; we make no claim of unbiasedness or of equivalence to full-chain back-propagation, and evaluate the truncation empirically.
Using more back-propagation steps consistently improves performance: the overall score increases from 53.16 at $K=1$ to 54.36, 55.05, and 55.38 at $K=2$, $3$, and $4$, while peak memory grows from 56 GB to 75 GB.
The marginal gains diminish, from $+1.20$ for $K{=}1\rightarrow2$ to $+0.69$ for $K{=}2\rightarrow3$ and $+0.33$ for $K{=}3\rightarrow4$, the last of which is close to the run-to-run standard deviation (\cref{sec:variance}).
This suggests that propagating reward gradients through more denoising steps provides stronger corrective supervision, with most of the benefit obtained within the last few steps; larger $K$ also incurs higher training cost and memory overhead, which limits further scaling in practice.

\subsection{Explanations: Length, Ordering, and Content}
\label{sec:explanation_ablation}
\begin{table}[htbp]
    \centering
    \setlength{\tabcolsep}{8pt}
    \scriptsize
    \caption{
        \textbf{Explanation ordering and content controls (VBench-2.0 Overall).}
        All variants use the same queries, data, and optimization settings and differ only in the supervised target sequence.
        The shuffled control keeps the answer tokens and the explanation length but pairs each query with an explanation drawn from another sample in the same supervision group with the same answer label.
        Three-seed statistics are reported where available (\cref{sec:variance}).
    }
    \vspace{1mm}
    \begin{tabular}{l|c|c}
        \toprule
        \textbf{Target format} & Single run & Three seeds (mean $\pm$ std) \\
        \midrule
        Answer only                                   & 53.30 & -- \\
        Shuffled explanation (length-matched)         & 52.47 & -- \\
        Explanation $\rightarrow$ Answer              & 55.24 & 55.31 $\pm$ 0.16 \\
        Answer $\rightarrow$ Explanation (default)    & {\bf 55.38} & {\bf 55.44 $\pm$ 0.26} \\
        \bottomrule
    \end{tabular}
    \label{tab:explanation_order}
\end{table}
\textbf{Explanation length.}
By default, we use reference explanations with an average length of 50 tokens.
We also adopt Chain-of-Thought (CoT) prompting to generate longer reference explanations (``Explanation w/ CoT'' in \cref{tab:quant-ablation-details}), which typically expands the explanation to over 100 words and introduces much more redundant content.
As shown in \cref{tab:additional_ablation}, CoT supervision can improve performance at the early stage of training, achieving 54.85 at 2K steps, which suggests that longer explanations initially provide richer diagnostic cues.
However, this benefit is short-lived: performance drops to 51.89 at 3K steps, indicating substantially earlier collapse than the non-CoT setting.
We attribute this to the excessive verbosity of CoT outputs, where many irrelevant tokens dilute useful gradients and amplify optimization noise during training.

\textbf{Explanation ordering.}
In our default target format the Yes/No token precedes the explanation, so the explanation is generated conditioned on an already fixed decision and cannot causally inform it.
We therefore describe the explanation as \emph{rationale supervision} rather than chain-of-thought reasoning: its role is to provide additional target tokens whose likelihoods, through the frozen VLM, are differentiated with respect to the generated video.
To test whether the order matters, we train a variant whose targets place the explanation before the answer, holding data, targets, and optimization fixed (\cref{tab:explanation_order}).
Explanation-first obtains 55.24 versus 55.38 for answer-first in a single run, and $55.31 \pm 0.16$ versus $55.44 \pm 0.26$ over three matched seeds; the paired mean difference of 0.13 points (95\% CI $[-0.12, 0.38]$) provides no evidence of an ordering effect.
This is expected, since both formats contain the same target tokens and ordering only changes the autoregressive conditioning, not the supervision content.
The meaningful contrast is between either explanation-containing format (55.31/55.44) and the answer-only objective (53.30).

\textbf{Explanation content.}
Longer targets alone could conceivably help simply by providing more tokens.
We rule this out with a length-matched control that shuffles explanations across samples within the same supervision group and answer label, so that each query is supervised with a fluent but unrelated explanation.
This control obtains 52.47, \emph{below} the answer-only objective (53.30) and below the base model, showing that the gain from explanations comes from their semantics rather than from the number of target tokens, and that mismatched rationales actively harm the reward signal.

\subsection{Role of the Reference Video}
\label{sec:reference_role}
\begin{table}[htbp]
    \centering
    \setlength{\tabcolsep}{8pt}
    \scriptsize
    \caption{
        \textbf{Role of the caption-matched reference video (VBench-2.0).}
        ``Shuffled reference'' pairs each caption with a real video from a different caption; ``SFT'' directly fine-tunes the generator on the same 5K real videos with a matched update budget.
        Diversity is the VBench-2.0 diversity score (higher means more cross-sample variation).
    }
    \vspace{1mm}
    \begin{tabular}{l|c|c}
        \toprule
        \textbf{Setting} & Overall & Diversity \\
        \midrule
        Base model                                & 52.99 & 58.6 \\
        PickScore~\cite{Pick-a-Pic}               & 49.62 & 29.5 \\
        VideoAlign~\cite{VideoAlign}              & 53.55 & 52.4 \\
        SFT on reference videos (matched budget)  & 53.42 & -- \\
        \methodname{}, no reference               & 53.15 & -- \\
        \methodname{}, shuffled reference         & 53.06 & -- \\
        \methodname{}, matched reference          & {\bf 55.38} & {\bf 64.4} \\
        \bottomrule
    \end{tabular}
    \label{tab:reference_role}
\end{table}
\begin{table}[htbp]
    \centering
    \setlength{\tabcolsep}{8pt}
    \scriptsize
    \caption{
        \textbf{Reference conditioning versus explanation supervision (VBench-2.0 Overall).}
        A $2{\times}2$ factorial over (i) whether Phy/VQ queries are conditioned on the caption-matched reference video and (ii) whether free-form explanations are supervised in addition to the Yes/No answers, on two training corpora.
        ``Answer only'' and ``Full'' use the same TA/VQ/Phy query set. The VideoUFO column uses 5,000 caption--video pairs from VideoUFO~\cite{VideoUFO} with the same filtering, query pipeline, TA:Phy:VQ ratio, initialization, and optimization settings.
        Rows within $\pm0.26$ of the base model (one run-to-run standard deviation, \cref{sec:variance}) are not distinguishable from it.
    }
    \vspace{1mm}
    \begin{tabular}{l|cc|cc}
        \toprule
        \textbf{Training setting} & Reference & Explanation & OpenVid-1M & VideoUFO \\
        \midrule
        Base model (Wan2.1-1.3B-T2V)  &            &            & 52.99 & 52.99 \\
        Answer only, no reference     &            &            & 53.02 & 53.10 \\
        Full feedback, no reference   &            & \checkmark & 53.15 & 53.22 \\
        Answer only, with reference   & \checkmark &            & 53.30 & 53.54 \\
        Full \methodname{}            & \checkmark & \checkmark & {\bf 55.38} & {\bf 55.15} \\
        \midrule
        \multicolumn{3}{l|}{Interaction (difference-in-differences)} & +1.95 & +1.49 \\
        \bottomrule
    \end{tabular}
    \label{tab:reference_factorial}
\end{table}
Phy and VQ queries condition the frozen critic on a caption-matched real video (\cref{sec:method:diffusion-drf}).
Since the generator is trained on on-policy samples and the reference receives no gradient, \methodname{} is not a form of supervised fine-tuning on real videos, but the reference could still bias the generator toward the reference distribution. We examine both questions.

\textbf{Matched references are required.}
\cref{tab:reference_role} compares the default model with a no-reference variant (Phy/VQ queries evaluate the generated video alone) and a shuffled-reference variant (each caption is paired with a real video from a different caption).
Removing references reduces VBench-2.0 Overall from 55.38 to 53.15 and shuffling them gives 53.06; both are within one run-to-run standard deviation of the base model (52.99, \cref{sec:variance}).
We therefore do not claim a measurable improvement without matched references: the gain of \methodname{} is contingent on reference-calibrated Phy/VQ feedback, and mere exposure to real videos is not sufficient.

\textbf{\methodname{} is not reference imitation.}
Directly fine-tuning the generator on the same 5K real videos with a matched update budget (SFT) obtains 53.42, well below \methodname{} (55.38), and \methodname{} \emph{increases} VBench-2.0 Diversity from 58.6 for the base model to 64.4, whereas the learned scalar rewards reduce it (52.4 for VideoAlign and 29.5 for PickScore, consistent with the prompt-unrequested stylization visible in \cref{fig:qual_compare_main,fig:single_prompt_ft,fig:additional_sp}).
Diversity measures cross-sample variation rather than style bias directly, so we use it only as supporting evidence.

\textbf{Reference conditioning and explanation supervision interact.}
To separate the contribution of references from that of the rich feedback, \cref{tab:reference_factorial} reports a complete $2{\times}2$ factorial over reference conditioning and explanation supervision, on OpenVid-1M and on an independently collected corpus of 5,000 VideoUFO~\cite{VideoUFO} caption--video pairs processed with the same filtering pipeline (OpenVid source overlaps and caption/visual near-duplicates removed) and identical query pipeline, TA:Phy:VQ ratio, initialization, and optimization.
Under answer-only supervision, references add only $+0.28$/$+0.44$ points (OpenVid/VideoUFO); without references, explanations add only $+0.13$/$+0.12$.
With full feedback, references add $+2.23$/$+1.93$, and with references fixed, explanations add $+2.08$/$+1.61$.
The difference-in-differences interaction is $+1.95$/$+1.49$, substantially larger than either isolated effect, on both corpora.
Neither component alone reproduces the full gain: references calibrate the Phy/VQ judgments, and the rich token-level feedback converts the resulting multi-dimensional judgments into gradients on on-policy videos.
Because the two corpora have different native captions we do not interpret their absolute-score difference as a controlled corpus effect; the matched within-corpus comparisons show that the interaction transfers to a second corpus.

\subsection{Loss Normalization}
\label{sec:loss_normalization}
For brevity the objective in \cref{sec:method:diffusion-drf} is written as a sum over queries and tokens.
The implementation normalizes it as a mean over valid target tokens,
\begin{equation}
\mathcal{L}_{\text{VQA}}
=
-\frac{\sum_{i=1}^{N}\sum_{t=1}^{L_i} m_{i,t}\,\log p_\phi\!\left(y_i^{(t)} \mid \mathbf{c}_i, y_i^{(<t)}\right)}
      {\sum_{i=1}^{N}\sum_{t=1}^{L_i} m_{i,t}},
\end{equation}
where $m_{i,t}\in\{0,1\}$ masks padding tokens, so that the effective learning rate does not scale with the number or length of the queries of a prompt.
Because prompts of different complexity yield different numbers of TA queries, we also evaluate a per-query mean (each query contributes equally) and a per-dimension-balanced mean (TA, Phy, and VQ contribute equally): they obtain 55.31 and 55.45 versus 55.38 for the valid-token mean.
The three schemes are within run-to-run variance of each other, showing that the gain is not driven by response-length or dimension-frequency weighting; we keep the valid-token mean in all reported results.

\subsection{Efficient Video Decoder}
\label{sec:h3ae}
The H3AE decoder~\cite{wu2025h3ae} used in the reward path (\cref{sec:exp:setup}) sits inside the differentiable chain, so we verify that it does not change the reward signal.
It is a lighter decoder trained to reproduce the reconstructions of the original Wan2.1 VAE decoder and is used solely to reduce memory; it is not a source of quality improvement.
At $K=2$, training with the original decoder obtains 54.7 VBench-2.0 Overall versus 54.4 with H3AE, a difference within one run-to-run standard deviation (\cref{sec:variance}), while peak memory increases from 63 GB to 81 GB.
H3AE reduces the memory of the decoding step by 24.3\%, which is what makes $K=4$ and $N_f=10$ fit on 80 GB GPUs.

\section{Limitations and Future Work}
\label{sec:limitations_future}
\textbf{Paired training data.}
Although \methodname{} needs neither a reward model nor preference annotations, its Phy and VQ dimensions require a caption-matched real video for every training prompt (\cref{sec:data_statistics}), and the gain is contingent on matched references (\cref{sec:reference_role}).
Domains without such a paired corpus would need one to be collected; extending the reference-free variant, for example with retrieved rather than paired references, is an open direction.

\textbf{Training--evaluation coverage.}
The model is tuned on 5K OpenVid-1M videos and evaluated on VBench-2.0 and VideoGen-Eval prompts. We verified that no training caption appears in the evaluation sets and that the reference/explanation interaction transfers to a second corpus (VideoUFO), but we have not characterized how far the evaluation prompts lie from the training distribution, and coverage of the training captions, rather than their number, likely bounds the improvement.

\textbf{Evaluator correlation.}
The critic is a VLM and both benchmarks contain learned evaluators. The human study (\cref{sec:human_protocol}) provides independent evidence against the base model only; human comparisons against the strongest reward-tuned baselines remain to be done.

\textbf{Sensitivity to the wording of cached explanations.}
The objective maximizes the likelihood of a single cached reference explanation. Natural language admits many equivalent phrasings, so a correct video whose content would elicit a differently worded explanation is penalized; caching makes comparisons reproducible but does not provide semantic invariance. Multi-paraphrase targets or semantic-matching objectives are natural remedies.

\textbf{Approximations in the differentiable path.}
Truncated back-propagation is a biased estimator whose bias we only characterize empirically (\cref{sec:truncation_ablation}), the H3AE decoder approximates the original VAE decoder (\cref{sec:h3ae}), and the critic sees $N_f=10$ uniformly sampled frames. The last choice is adequate for 49-frame clips but may miss multi-stage actions or delayed physical effects in longer videos, so long-horizon scalability is untested.

\textbf{Critic scale and run-to-run variance.}
Our critics are capped at 7B parameters; the monotonic trend from 1B to 3B to 7B (53.10, 53.72, 55.38) is established only within this range. Three-seed variance is reported for our method (\cref{sec:variance}) but the baselines are single runs, so small margins between baselines should not be over-interpreted.

\textbf{Training stability and cost.}
Video-generation alignment tuning remains challenging. Prior reward-tuning approaches~\cite{zheng2026sagegrpo,DanceGRPO,VideoAlign} can empirically be trained for fewer than 200 steps before collapsing, while \methodname{}, to the best of our knowledge, is the first to remain stable up to 3K steps, although it still collapses afterward. A promising direction is to incorporate stabilization strategies from GRPO-style optimization, such as reward clipping, to further improve long-horizon training stability.
Alignment tuning also requires a full-trajectory rollout of 25 denoising steps per training sample, and the longer schedule that our stability permits means that the gain over VideoAlign comes at roughly 3.2$\times$ its GPU-hours (\cref{sec:training_cost}). Future work may explore step-distilled generators and quantized rollout (INT4/INT8)~\cite{yin2025slowbidirectionalfastautoregressive} to further accelerate reward tuning for video generation.

\section{Broader Impacts}
\label{sec:broader_impacts}
\methodname{} improves the prompt adherence, physical plausibility, and visual quality of text-to-video generators without collecting human preference data, which lowers the cost of aligning generators to new requirements.
The same property can lower the barrier to producing realistic but misleading video content.
Our method does not add capabilities for identity synthesis or for removing provenance signals, but it also does not add safeguards of its own; fine-tuned checkpoints should be released under the base model's license and usage policy and, where possible, accompanied by safeguards such as content filters and watermarking.
Because the critic is a frozen VLM, its biases (for example, toward particular visual styles or demographics) can be transferred to the generator; the reference-guided Phy/VQ queries and the human study in \cref{sec:human_protocol} only partially control for this, and auditing the critic's judgments for such biases is an important direction.
The human study collected anonymized pairwise preferences over generated videos and no personal data about the annotators.

\section{Additional Visual Results}
\label{sec:visual_results}

\begin{figure*}[t]
\vspace{-2mm}
\centering
\includegraphics[width=0.9 \linewidth]{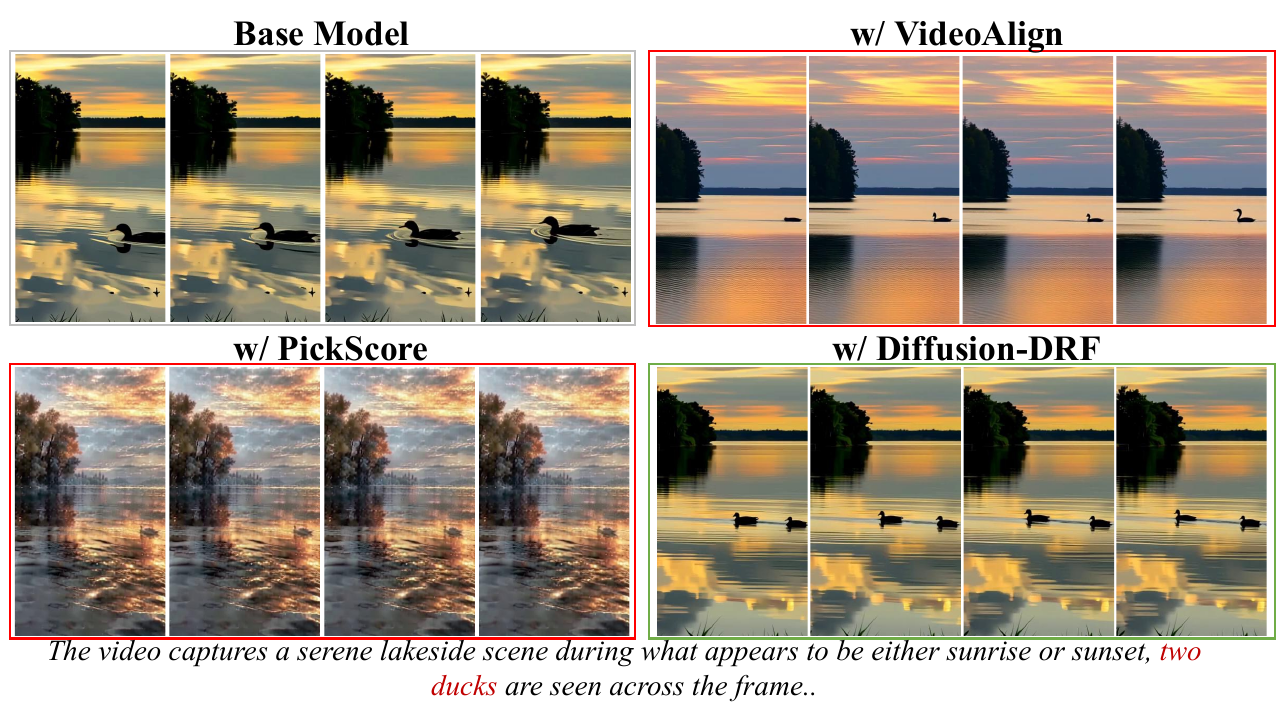}
\includegraphics[width=0.9 \linewidth]{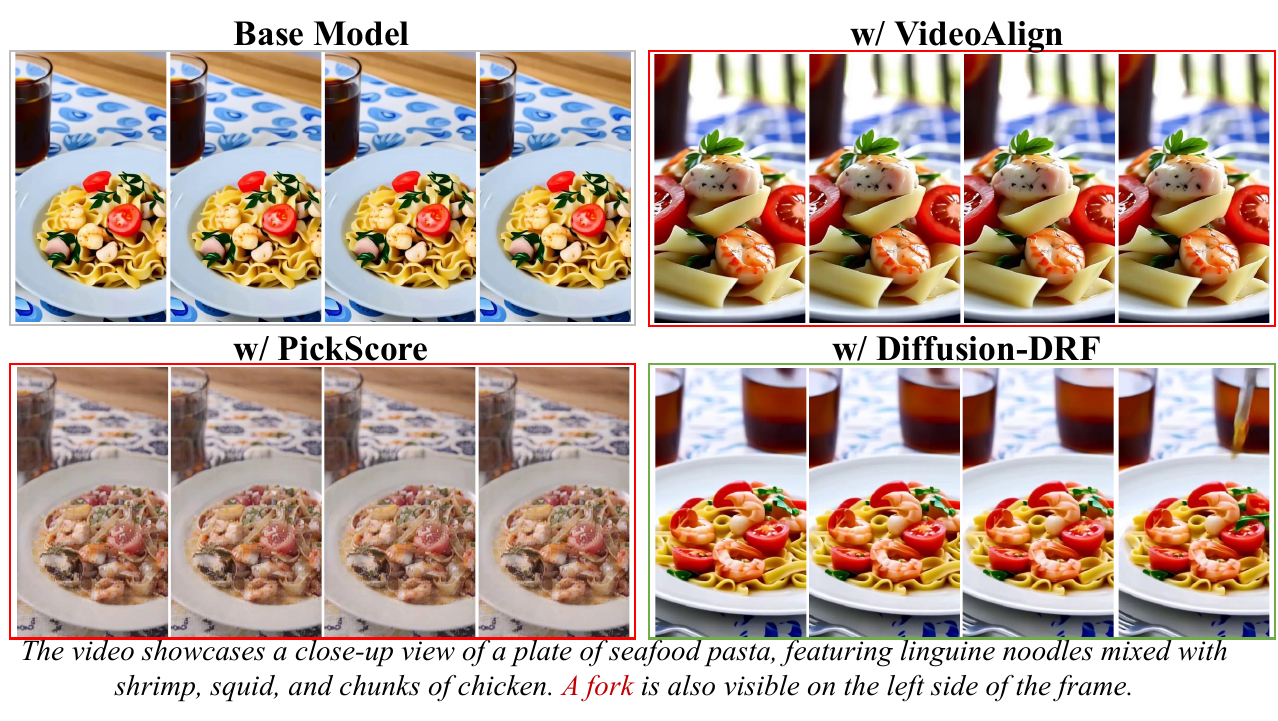}

\caption{
\textbf{Additional Samples of Single Prompt Finetuning.}
}
\label{fig:additional_sp}
\end{figure*}
\begin{figure*}[t]
\vspace{-2mm}
\centering
\includegraphics[width=0.9 \linewidth]{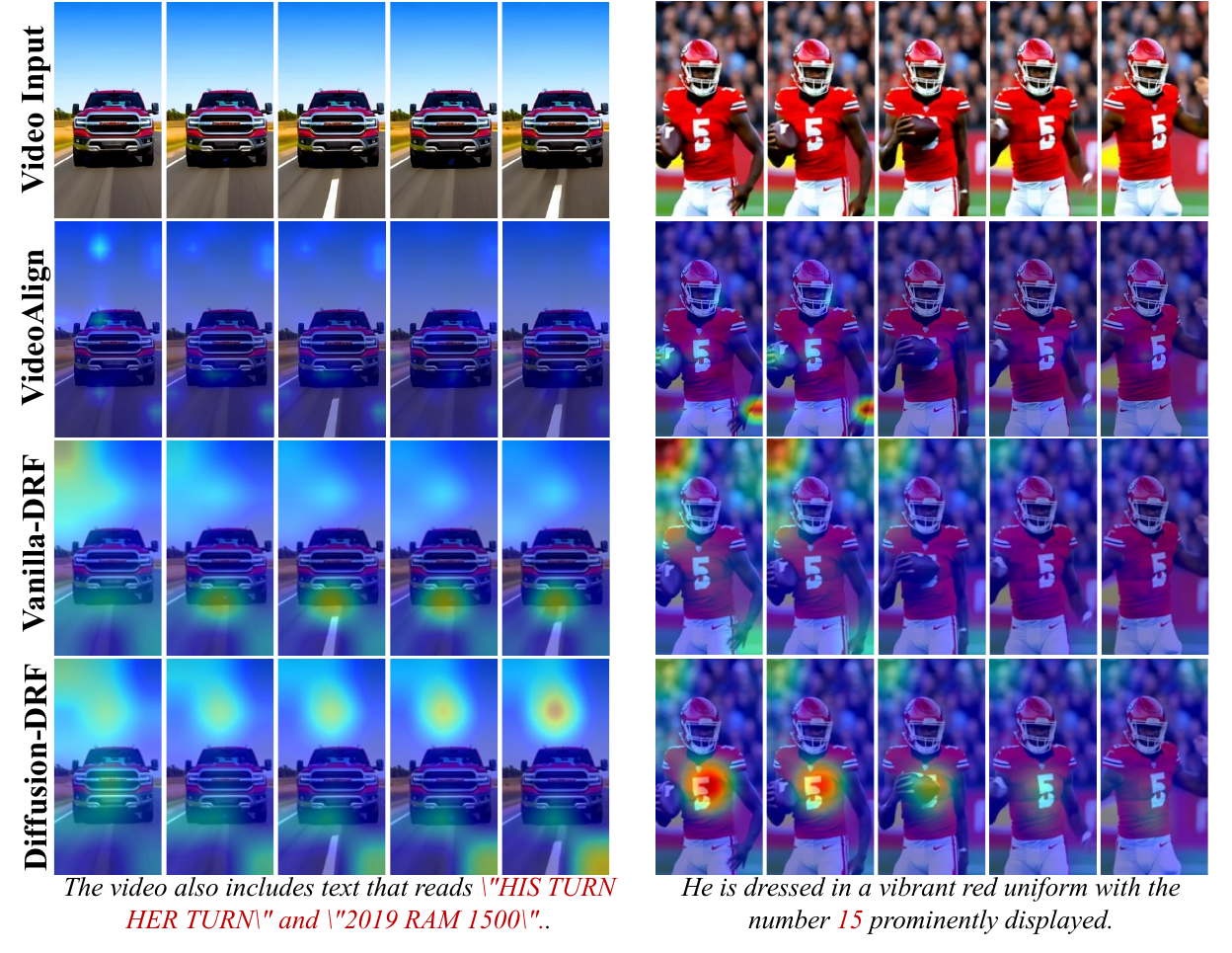}
\includegraphics[width=0.9 \linewidth]{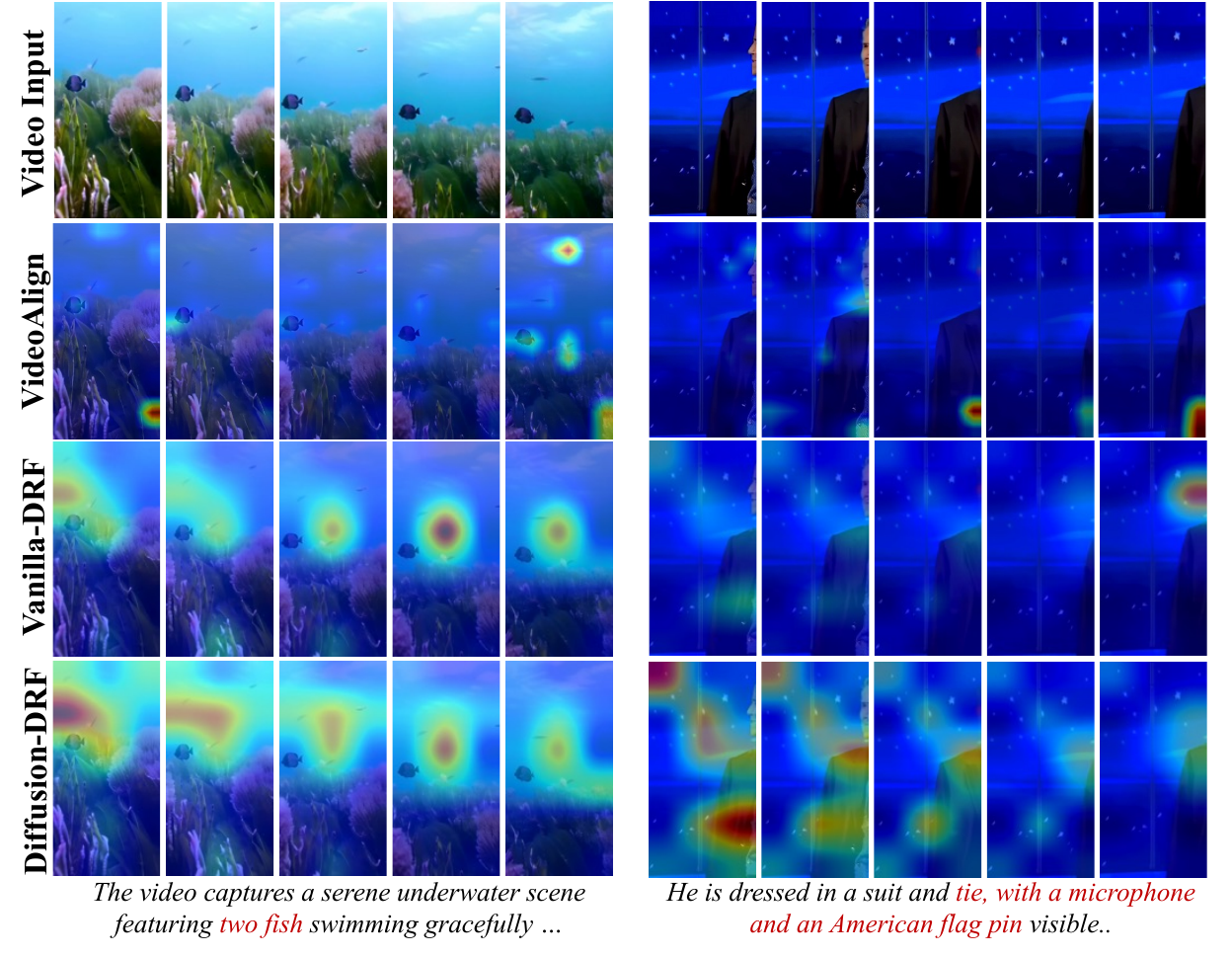}

\caption{
\textbf{Additional Gradient Heatmaps - Part1.}
}
\label{fig:additional_heatmaps1}
\end{figure*}

\begin{figure*}[t]
\vspace{-2mm}
\centering
\includegraphics[width=0.9 \linewidth]{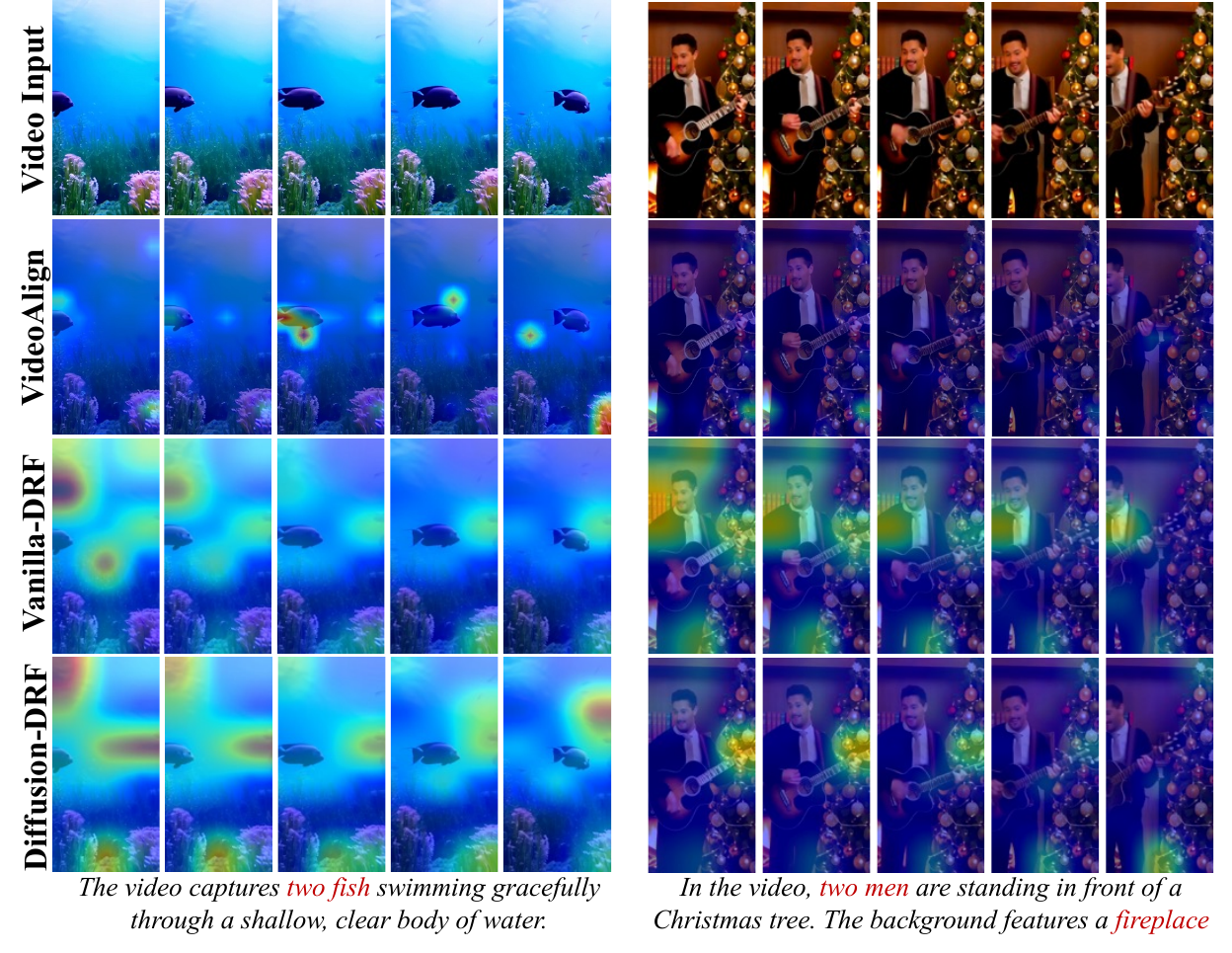}
\includegraphics[width=0.9 \linewidth]{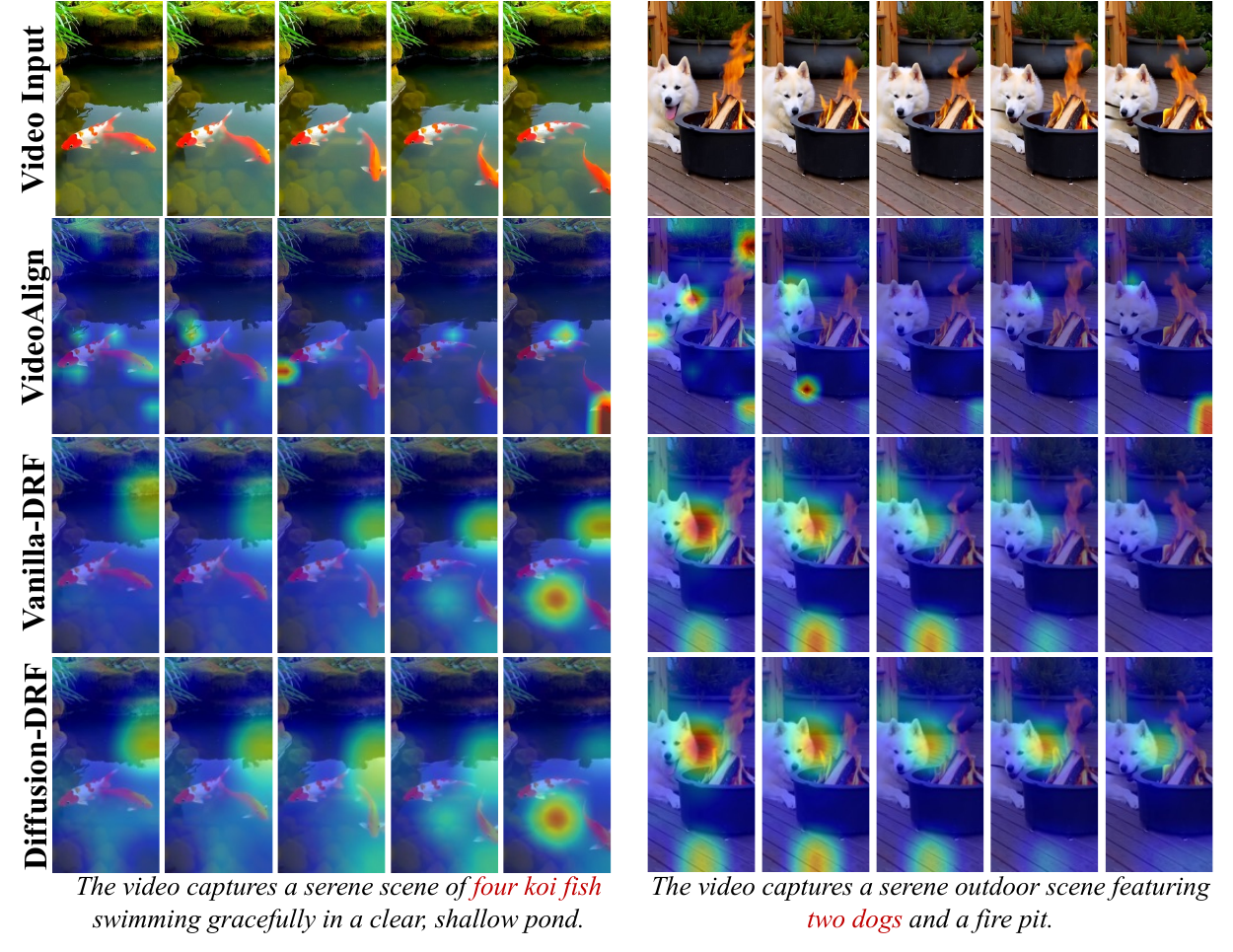}

\caption{
\textbf{Additional Gradient Heatmaps - Part2.}
}
\label{fig:additional_heatmaps2}
\end{figure*}
\cref{fig:additional_sp} shows more examples of single prompt fine-tuning.
\cref{fig:additional_heatmaps1,fig:additional_heatmaps2} show further visualization of the gradient heatmaps.
We also provide visual results in video format on the project page.
On the project page, we first present qualitative comparisons among the base model, the model fine-tuned with VideoAlign, and our \methodname{} model.
These visual results show that our method improves both text-video alignment and physical fidelity.
We also provide examples demonstrating that when the base model already produces high-quality videos, our method preserves these strengths.
In contrast, VideoAlign may lead to over-optimization and deteriorate originally high-quality generations.
We additionally present single prompt fine-tuning videos and several cases in which our method performs particularly well.

\section{Detailed Prompts}
\label{sec:detailed_prompts}
In this section, we present the prompts used in our prompting pipeline (\cref{fig:prompt_decompose_p1,fig:prompt_decompose_p2}) as well as the prompts used during training (\cref{fig:ta_prompts,fig:phy_prompts,fig:vq_prompts}), which cover the various facets used for analyzing the generated videos.

\begin{figure*}
    \centering
\begin{tcolorbox}[
  colback=black!5!white,
]
{\bf Prompts of caption decomposition - Part1:}

Please analyze the following video generation prompt by breaking it down based on the following key components: \\
1. Environment: Describe the overall setting and static elements of the environment. \\
2. Object(s): List the main objects/entities in the scene. \\
3. Objects' Motion: Describe how each object moves or interacts within the environment. Include direction, speed, and behavior. \\
4. Object Location / Spatial Distribution: Describe the spatial layout of objects in the scene. Where are they located (e.g., left/right/center/far/near)? Is the composition symmetrical or unbalanced? \\
5. Color Requirement: Describe any mentioned or implied colors for objects or environment. Mention the dominant tone or palette. \\
6. Lighting: Describe the light source(s), brightness, shadows, and general mood (e.g., backlit, dim, dramatic, diffuse sunlight). \\
7. Letter/Text Presence: Indicate whether there are any textual elements (e.g., signs, billboards), whether text is legible, and how it integrates into the scene. \\
8. Camera Motion: Describe how the camera moves or stays still (e.g., tracking, panning, zooming, handheld, fixed). Also indicate how this affects perception of the scene.
Return the result in a structured bullet-point format, your response of each elements should be as simplified as possible. \\

Important:
Only extract the  the element is merely described atmospherically or implicitly observed — only count it if there is a clear instruction or strong implication to generate it.

Here is the prompt to analyze:

\textcolor{blue}{\{video caption\}}

The answer format should be in dict format: \\
**{``Environment": content, ``Object(s)": content, ``Object Location/Spatial Distribution": content, ``Objects' Motion": content, ``Color Requirement": content, ``Camera Motion": content}**
\\
Here is a decomposed example for you:\\
The video generation prompt is \\
``a moment on a rainy day in a city. The street, slick with rain, reflects the surrounding buildings and trees, creating a mirror-like surface. Two motorbikes, one blue and the other white, are making their way down this wet road. They are moving towards the camera, their tires kicking up droplets of water.  On the left side of the street, several tents and umbrellas have been set up, providing shelter from the rain. These structures add a splash of color to the otherwise gray scene. On the right side of the street, a red and yellow sign stands out, although the text on it is not visible.  The sky overhead is a blanket of gray, heavy with rain clouds. Despite the inclement weather, there's a certain tranquility to the scene. It's as if time has slowed down, allowing one to fully take in the details of this rainy day in the city." \\

\end{tcolorbox}

\caption{Prompts used for the caption decomposition. Part 1}
\label{fig:prompt_decompose_p1}
\end{figure*}

\begin{figure*}
    \centering
\begin{tcolorbox}[
  colback=black!5!white,
]
{\bf Prompts of caption decomposition - Part2:}

Your answer: \\
\{\\
`Environment': `1). A city street during a rainy day. 2). The road is slick with rain, creating a mirror-like reflection of surrounding buildings and trees. 3). The sky is gray and heavy with rain clouds.' \\

`Object(s)': `1). Two motorbikes. 2). Several tents and umbrellas. 3). A red and yellow sign (text not visible).' \\
`Objects' Motion': `1). The two motorbikes are moving toward the camera. 2). Their tires are kicking up water droplets from the wet road. 3). Other objects (tents, umbrellas, sign) are static.' \\
`Object Location / Spatial Distribution': `1).The motorbikes are on the street. 2). Tents and umbrellas are placed on the left side of the street. 3). The sign is on the right side.' \\
`Color Requirement': `1). The overall palette is gray, reflecting the rainy weather. 2). The motorbikes are blue and white. 3). The sign is red and yellow. 4). The tents and umbrellas are colorful.' \\
`Lighting:': `Not explicitly stated.'\\
`Letter/Text Presence': `Not explicitly stated.'\\
`Camera Motion': `Not explicitly stated.'\\
\}
\end{tcolorbox}

\caption{Prompts used for the caption decomposition - Part2.}
\label{fig:prompt_decompose_p2}
\end{figure*}
\begin{figure*}
    \centering
\begin{tcolorbox}[
  colback=black!5!white,
]

{\bf Prompts of TA:}

Given this AI-generated video, does it successfully fulfill the \textcolor{blue}{\{key\}} condition: \textcolor{blue}{\{description\}}? \\
Respond with 'Yes' or 'No', Answer 'Yes' if the video largely matches the description. Answer 'No' if the video clearly contradicts the description. The presence of additional elements in the video is acceptable as long as they do not conflict with the core description. Please provide a brief explanation for your answer. \\
Provide your analysis and explanation in JSON with keys: answer (e.g., Yes or No), explanation.

\end{tcolorbox}

\caption{Prompts used in the training for text-video alignment (TA).}
\label{fig:ta_prompts}
\end{figure*}

\begin{figure*}
    \centering
\begin{tcolorbox}[
  colback=black!5!white,
]

{\bf Prompts of Phy:}

You are a careful video forensics assistant. You are given two videos: a test video (the first 10 frames) which is ai-generated and a real video (the last 10 frames). The test video is generated by the caption of the real video. The caption is 

\textcolor{blue}{\{video prompt\}}

Your task is evaluating whether the test video shows physics-related defects. You should compared both videos and use the provided caption as high-level intent. Focus on physical plausibility, not style or aesthetics.

You need to analyze it in these aspects:

- Liquid flow irregularity – e.g., non-inertial or discontinuous flow, volume popping, gravity-inconsistent motion, impossible splashes.

- Abnormal object deformation – e.g., rigid objects bending/stretching without cause, topology changes (parts merging/splitting).

- Abnormal texture/material change – e.g., surface turns matte$\rightarrow$glossy without cause, texture swimming/flicker detached from geometry.

- Abnormal motion – e.g., inertia/acceleration violations, teleporting, time reversals, jitter not explained by camera motion.

- Unnatural interpenetration – e.g., objects passing through each other or ground, missing collisions/contacts.

If the prompt doesn't have related physical aspects, you should answer `No'.

Output strictly as JSON with this schema (no extra text):

\{ \\
    ``liquid flow irregularity": ``Yes or No",\\
    ``abnormal deformation":  ``Yes or No",\\
    ``abnormal texture change":    ``Yes or No",\\  
    ``abnormal motion":   ``Yes or No",  \\
    ``unnatural interpenetration":  ``Yes or No", \\
\}

\end{tcolorbox}

\caption{Prompts used in the training for physical fidelity (Phy).}
\label{fig:phy_prompts}
\end{figure*}

\begin{figure*}
    \centering
\begin{tcolorbox}[
  colback=black!5!white,
]

{\bf Prompts of VQ:}

Compare the test video (the first 10 frames) to the reference frame (the last 10 frames) and decide if the video shows obvious visual-quality (VQ) defects relative to the reference.\\
Consider only: blur (defocus/motion), compression artifacts (blocking/ringing/mosquito), noise/grain, banding, flicker, rolling-shutter, aliasing/moire, over-smoothing.

Return ONLY this JSON (no timestamps):\\
\{ \\
  ``has obvious defect": Yes / No,\\
  ``dominant issue": ``none / defocus blur / motion blur / blocking / ringing / mosquito noise / grain noise / banding / flicker / rolling shutter / aliasing / moire / over or smoothing", \\
  ``evidence": [``short visual cues vs reference, e.g., softer edges than reference', block edges visible around text'"],\\
\} \\
Rules:

- Compare against the reference frame’s look (sharpness, texture, edges, tones).

- Be conservative; if unsure, choose false and note 'uncertain' in evidence.'''
 
\end{tcolorbox}

\caption{Prompts used in the training for visual quality (VQ).}
\label{fig:vq_prompts}
\end{figure*}

\end{document}